\documentclass[11pt]{article}

\usepackage[final]{acl}

\usepackage{times}
\usepackage{latexsym}

\usepackage[T1]{fontenc}
\usepackage[utf8]{inputenc}

\usepackage{microtype}

\usepackage{inconsolata}

\usepackage{graphicx}

\usepackage[utf8]{inputenc}
\usepackage{hyperref}
\usepackage[most]{tcolorbox}
\usepackage{xspace}
\usepackage{enumitem}
\usepackage{subcaption} 
\usepackage{float}
\usepackage{wrapfig}
\usepackage{xcolor}
\usepackage{colortbl}
\usepackage{makecell}
\usepackage{appendix}
\usepackage{caption}
\usepackage{multirow}
\usepackage{colortbl}
\usepackage{graphicx}
\usepackage[ruled,linesnumbered]{algorithm2e}
\usepackage{multicol}
\usepackage{booktabs}
\usepackage{amssymb}
\usepackage{pifont}
\usepackage{tabularx}
\usepackage{longtable}
\usepackage{array}
\usepackage{listings}
\usepackage{xcolor}
\tcbuselibrary{skins,breakable}
\usepackage{caption}
\usepackage{subcaption}

\newcommand{\hang}[1]{}

\definecolor{mygray}{gray}{.88}
\definecolor{mycyan}{cmyk}{.15,0,0,0}
\definecolor{mycyan2}{cmyk}{.85,0,0,0}
\definecolor{mygreen}{rgb}{0.19, 0.79, 0.02}
\definecolor{midnightgreen}{rgb}{0.0, 0.29, 0.33}
\definecolor{darkgreen}{RGB}{0,160,0}

\definecolor{mygray}{gray}{.88}
\definecolor{mycyan}{cmyk}{.15,0,0,0}
\definecolor{mycyan2}{cmyk}{.85,0,0,0}
\definecolor{mygreen}{rgb}{0.19, 0.79, 0.02}
\definecolor{midnightgreen}{rgb}{0.0, 0.29, 0.33}
\definecolor{darkgreen}{RGB}{0,160,0}
\definecolor{backyellow_soft}{rgb}{1.0,1.0,0.8}
\definecolor{backred}{RGB}{255, 190, 190}
\definecolor{backblue}{RGB}{210, 230, 250}

\newcommand{\bench}{\textsc{VistaHop}\xspace}

\newcommand{\eval}{\textsc{VistaArena}\xspace}

\newcommand{\notcheckmark}{\textcolor{black}{\bcmark\kern-1.1ex\raisebox{.7ex}{\rotatebox[origin=c]{125}{--}}}\color{black}}
\newcommand{\bcmark}{\color{blue}{\ding{51}}}
\newcommand{\cmark}{\color{darkgreen}{\ding{51}}}
\newcommand{\xmark}{\color{red}{\ding{55}}}

\definecolor{todocolor}{rgb}{0.9,0.1,0.1}

\newcommand{\eg}{\hbox{\emph{e.g.}}\xspace}

\usepackage{pythonhighlight}

\definecolor{codegreen}{rgb}{0,0.6,0}
\definecolor{codegray}{rgb}{0.5,0.5,0.5}
\definecolor{codepurple}{rgb}{0.58,0,0.82}
\definecolor{backcolour}{rgb}{0.97,0.97,0.95}
\definecolor{forestgreen}{rgb}{0.28,0.62,0.37}

\lstdefinestyle{mystyle}{
    backgroundcolor=\color{backcolour},   
    commentstyle=\color{codegray},
    keywordstyle=\color{codepurple},
    numberstyle=\tiny\color{codegray},
    stringstyle=\color{blue},
    basicstyle=\ttfamily\footnotesize,
    breakatwhitespace=false,         
    breaklines=true,                 
    captionpos=b,                    
    keepspaces=true,                 
    numbers=left,                    
    numbersep=5pt,                  
    showspaces=false,                
    showstringspaces=false,
    showtabs=false,                  
    tabsize=4,
}

\title{
  VistaHop: Benchmarking Long-Horizon Visual DeepSearch
}
\author{
  Hang He$^{1,2,4}$, Chuhuai Yue$^{2}$, Chengqi Dong$^{3,2}$,
  Chengcheng Wan$^{1,4,\dagger}$, Ting Su$^{1}$, Haiying Sun$^{1}$, \\
  \textbf{Jiajun Chai}$^{2}$, \textbf{Xiaohan Wang}$^{2}$, 
  \textbf{Guojun Yin}$^{2,\dagger}$ \\
  $^{1}$East China Normal University\quad
  $^{2}$Meituan\\
  $^{3}$University of Science and Technology of China\\
  $^{4}$Shanghai Innovation Institute\\[2pt]
  \href{https://visualdeepsearch.github.io/}{%
    \raisebox{-0.12em}{\includegraphics[height=1em]{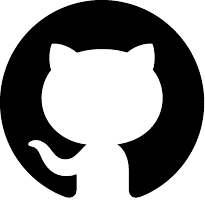}}%
    \ \textbf{Project Page}}
}

\begin{document}
\maketitle

\begingroup
\renewcommand{\thefootnote}{}
\footnotetext{$^{\dagger}$ Corresponding authors.}
\endgroup

\begin{abstract}
Visual DeepSearch tasks require multimodal large language models (MLLMs) to resolve complex visual queries by repeatedly inspecting image regions, grounding reasoning in visual evidence, and connecting fine-grained clues across multiple steps. However, existing benchmarks primarily evaluate single-step visual understanding or isolated visual-query response generation. They have limited difficulty, limited search horizons, and single-pass image inspection, and thus fail to evaluate models' ability to iteratively revisit visual evidence and reason across multiple steps.

In this work, we introduce \bench, a benchmark designed specifically to evaluate Visual DeepSearch. It evaluates repeated image inspection, visual-anchor grounding, and long-horizon evidence traversal across different visual regions. \bench comprises 600 images, 25 visual search scenarios, and 600 Visual DeepSearch tasks. We also propose \eval, a unified evaluation framework that supports tool-based interactions, including visual retrieval, image inspection, and evidence-grounded reasoning. Experiments show that even state-of-the-art MLLMs remain far from solving \bench, with the best-performing model, SenseNova-MARS-32B, achieving only 26.33\% Pass@1. These findings highlight the importance of specialized benchmarks and improved agentic methods for Visual DeepSearch.
\end{abstract}

\section{Introduction}
\sloppy

\begin{figure}[!t]
    \centering
    \begin{subfigure}[b]{\columnwidth}
        \centering
        \includegraphics[width=0.8\textwidth]{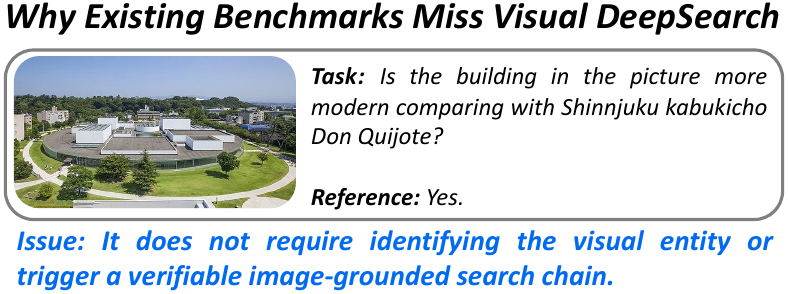}
        \vspace{-5pt}
        \caption{Non-vision-centric search.}
        \label{fig:intro_text_dominant}
    \end{subfigure}
    
    \begin{subfigure}[b]{\columnwidth}
        \centering
        \includegraphics[width=0.83\textwidth]{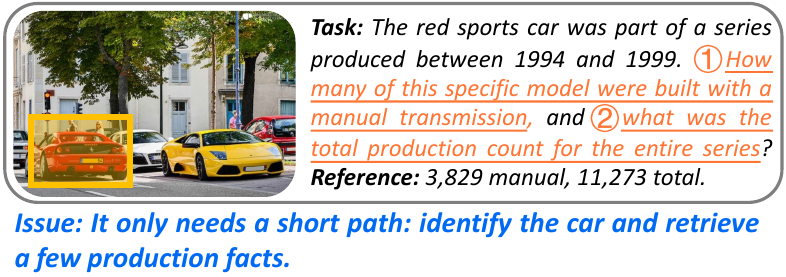}
        \vspace{-5pt}
        \caption{Limited search horizon.}
        \label{fig:intro_shallow_search}
    \end{subfigure}
    
    \begin{subfigure}[b]{\columnwidth}
        \centering
        \includegraphics[width=0.83\textwidth]{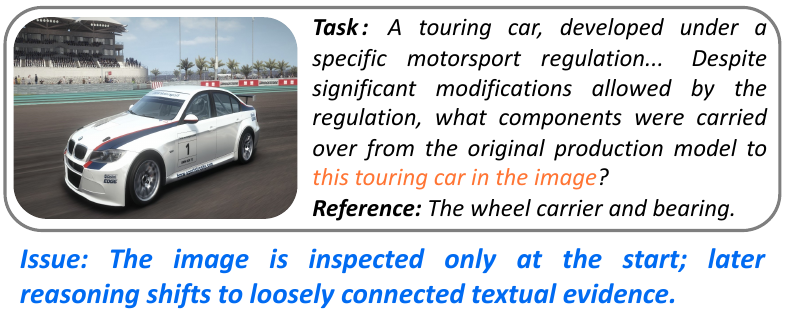}
        \vspace{-5pt}
        \caption{No repeated image inspection.}
        \label{fig:intro_one_shot}
    \end{subfigure}

    \begin{subfigure}[b]{\columnwidth}
        \centering
        \includegraphics[width=0.83\textwidth]{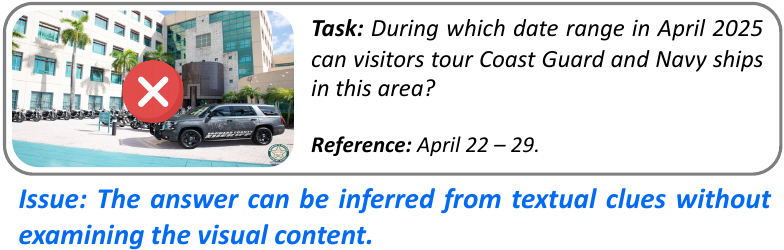}
        \vspace{-5pt}
        \caption{Solvable without image inspection.}
        \label{fig:intro_leakage}
    \end{subfigure}
    \vspace{-10pt}
    \caption{The four limitations of released benchmarks~\cite{jiang2025mmsearch,zeng2026vdrbench,geng2026webwatcher}: non-vision-centric search, limited search horizon, no repeated image inspection, and tasks solvable without image inspection. All images and queries come from the original benchmarks.}
    \label{fig:intro}
    \vspace{-10pt}
\end{figure}

Recent advances in multimodal large language models (MLLMs) have enabled agents to move beyond passive image understanding toward fine-grained visual perception, region-level grounding, and tool-assisted multimodal reasoning~\cite{dong2025insightv,li2025dyfo,bigverdi2025perception,qi2025cogcom,li2025vocot,he2026finer1,chai2025rlfactory,dong2026trainingmultiimagevisionagents}. 
Together with recent multimodal search and agentic reasoning systems~\cite{jiang2025mmsearch,geng2026webwatcher,tao2026mmsearchplus,he2026localsearchbenchbenchmarkingagenticsearch,
chen2025toolforgedatasynthesispipeline}, these capabilities support \textbf{Visual DeepSearch}, a vision-centric, long-horizon agentic search setting in which agents iteratively inspect task-relevant image regions, retrieve and verify external information, and connect image-grounded clues through multi-step evidence-seeking trajectories~\cite{li2025dyfo,dong2025insightv}. 
Unlike text-dominant agentic search, Visual DeepSearch places repeated visual evidence seeking and region-level grounding at the center of the reasoning process.

However, as illustrated in Figure~\ref{fig:intro}, existing visual benchmarks are insufficient for evaluating Visual DeepSearch. \emph{First}, most involve non-vision-centric search: they rely on static visual-query instances and rarely require fine-grained visual retrieval, region-level inspection, or evidence revisiting. Visual perception is often front-loaded, used only initially while reasoning proceeds via language without revisiting visual evidence. \emph{Second}, their tasks tend to have limited search horizons, allowing direct responses rather than multi-step evidence chains. \emph{Third}, some benchmark tasks are solvable without image inspection, as their targets can be inferred from textual cues, prior knowledge, or memorized associations without genuine visual search. \emph{Finally}, many benchmarks rely heavily on manual task construction and do not fully release their construction pipelines, making them costly to scale and limiting reproducibility, quality verification, and extension. These limitations make it difficult to systematically evaluate whether current MLLMs can support deep, iterative, and evidence-grounded visual search. 

To bridge these gaps, we introduce \textbf{\bench}, a benchmark designed specifically to evaluate Visual DeepSearch. 
\bench evaluates whether models can repeatedly inspect images, identify visual anchors, and connect image-grounded evidence with external knowledge through long-horizon evidence chains. 
It contains 600 Visual DeepSearch tasks, with 74.3\% categorized as L3 tasks that require at least 10 evidence steps. 

We further develop \textbf{\eval}, a unified evaluation framework for MLLMs under direct response generation, search-augmented reasoning, and multi-anchor reasoning settings. 
\eval enables systematic analysis of response correctness, visual grounding, evidence revisiting, and cross-chain reasoning in Visual DeepSearch. Across seven representative MLLMs, the best-performing model achieves only 26.33\% Pass@1.

In summary, our contributions are threefold.
% \vspace{-3pt}
\begin{itemize}[leftmargin=*, itemsep=0pt, topsep=2pt, parsep=0pt]
    \item We introduce \textbf{\bench}, a benchmark for evaluating long-horizon evidence traversal, repeated image inspection, and evidence-grounded response generation through multi-chain Visual DeepSearch tasks.

    \item We design an automated and scalable construction process that generates visually grounded, long-horizon reasoning tasks while controlling data quality and reducing text-only shortcuts.

    \item We develop \textbf{\eval}, a unified evaluation framework for MLLMs, and show that current models remain limited in long-horizon evidence traversal, visual evidence revisiting, and multi-anchor evidence fusion.
\end{itemize}

% \section{Method}
% \input{sections/2-background.tex}
\section{Related Work}
\sloppy
\begin{table*}[t]
\centering
\scriptsize
\caption{Feature audit of existing benchmarks for evaluating \textit{Visual DeepSearch} capabilities. ``{\cmark}'' indicates fully addressed, ``{\notcheckmark}'' indicates partially addressed, and ``{\xmark}'' indicates not addressed; see Appendix~\ref{appendix:benchmark_feature_audit} for the operational criteria.}
\label{tab:benchmark_comparison}
\setlength{\tabcolsep}{0.7\tabcolsep}
\resizebox{\textwidth}{!}{
\begin{tabular}{l c c c c c c}
\toprule[1.5pt]
\textbf{Benchmark} 
& \textbf{Vision-}
& \textbf{Fine-grained}
& \textbf{Repeated}
& \textbf{Deep}
& \textbf{Temporal}
& \textbf{Target} \\
% & \textbf{Process} \\
% & \textbf{Open} \\
& \textbf{centric search}
& \textbf{visual retrieval}
& \textbf{image inspection}
& \textbf{long-horizon search}
& \textbf{validity}
& \textbf{uniqueness} \\
% & \textbf{evaluation} \\
% & \textbf{pipeline} \\
\midrule
MMSearch~\cite{jiang2025mmsearch}
& \notcheckmark
& \notcheckmark
& \xmark
& \xmark
& \xmark
& \notcheckmark \\
% & \notcheckmark \\
% & \notcheckmark \\

MMSearch-Plus~\cite{tao2026mmsearchplus}
& \cmark
& \cmark
& \cmark
& \notcheckmark
& \notcheckmark
& \notcheckmark \\
% & \cmark \\
% & \notcheckmark \\

VDR-Bench~\cite{zeng2026vdrbench}
& \cmark
& \cmark
& \cmark
& \xmark
& \xmark
& \cmark \\
% & \notcheckmark \\
% & \notcheckmark \\

BrowseComp-$V^3$~\cite{zhang2026browsecompv3}
& \notcheckmark
& \notcheckmark
& \notcheckmark
& \notcheckmark
& \notcheckmark
& \cmark \\
% & \cmark \\
% & \cmark \\

MMDeepResearch-Bench~\cite{huang2026mmdeepresearch}
& \notcheckmark
& \notcheckmark
& \xmark
& \notcheckmark
& \xmark
& \notcheckmark \\
% & \cmark \\
% & \notcheckmark \\

VTC-Bench~\cite{zhu2026vtcbench}
& \notcheckmark
& \xmark
& \cmark
& \xmark
& \xmark
& \cmark \\
% & \cmark \\
% & \cmark \\

AgentVista~\cite{su2026agentvista}
& \notcheckmark
& \notcheckmark
& \notcheckmark
& \notcheckmark
& \notcheckmark
& \notcheckmark \\
% & \notcheckmark \\
% & \notcheckmark \\

ARK~\cite{lin2026ark}
& \notcheckmark
& \cmark
& \xmark
& \xmark
& \xmark
& \notcheckmark \\
% & \xmark \\
% & \notcheckmark \\

DeepWideSearch~\cite{lan2025deepwidesearch}
& \xmark
& \xmark
& \xmark
& \notcheckmark
& \notcheckmark
& \notcheckmark \\
% & \notcheckmark \\
% & \cmark \\

MTA-Agent~\cite{peng2026mtaagent}
& \cmark
& \notcheckmark
& \notcheckmark
& \notcheckmark
& \xmark
& \cmark \\
% & \cmark \\
% & \cmark \\

OpenSearch-VL~\cite{chen2026opensearchvl}
& \cmark
& \cmark
& \cmark
& \notcheckmark
& \xmark
& \notcheckmark \\
% & \cmark \\
% & \cmark \\

\midrule
\textbf{\bench}
& \cmark
& \cmark
& \cmark
& \cmark
& \cmark
& \cmark \\
% & \notcheckmark \\
% & \cmark \\
\bottomrule[1.5pt]
\end{tabular}
}
\vspace{-5pt}
\end{table*}

% 出现一题多解
\subsection{Visual DeepSearch}
Recent multimodal search and agentic reasoning studies~\cite{du2026longhorizon,geng2026deepsearchworld,dong2026s1deepresearch,chen2025pass,li2026reverse,zhang2026vsearcher,guo2026e3tirenhancedexperienceexploitation,guo2026agentreinforcementlearningpivotalaware}, such as MMSearch~\cite{jiang2025mmsearch}, MMSearch-R1~\cite{wu2025mmsearchr1}, DeepMMSearch-R1~\cite{narayan2025deepmmsearchr1}, Vision-DeepResearch~\cite{huang2026visiondeepresearch}, WebWatcher~\cite{geng2026webwatcher}, MTA-Agent~\cite{peng2026mtaagent}, OpenSearch-VL~\cite{chen2026opensearchvl}, Visual-Seeker~\cite{zhang2026visualseeker}, SimpleSearch-VL~\cite{dai2026simplesearchvl}, SearchEyes~\cite{jiao2026searcheyes}, ProMMSearchAgent~\cite{yan2026prommsearchagent,zhao2025redone}, HyperEyes~\cite{li2026hypereyes}, Agent-X~\cite{ashraf2025agentx}, M$^3$Searcher~\cite{yu2026m3searcher}, MC-Search~\cite{ning2026mcsearch,liu2026automated}, AndroTMem~\cite{shi2026androtmem}, and CirrusBench~\cite{yu2026cirrusbench,zhou2026look,zhou2026one} show a clear shift from passive image understanding to active visual evidence seeking.

This emerging setting places visual evidence collection, region-level grounding, and long-horizon evidence traversal at the center of the search process. 
Recent region-level and tool-augmented visual reasoning methods~\cite{zhong2025focus,shi2026sieve,sarch2025vigorl}, including VLM-R$^3$~\cite{jiang2025vlmr3}, Chain-of-Focus~\cite{zhang2025chainoffocus,zhu2026medeyes}, TikArt~\cite{ding2026tikart}, ToolsRL~\cite{dong2026toolsrl,li2026reasoningtoolusecompeteagentic,liang2025ropo,liang2026boosting,cai2025reinforcement,cai2026vi,liu2026socraticpo,wang2025soay}, Pixelis~\cite{zhou2026pixelis}, CodeV~\cite{hou2025codev,liang2025boosting,chen2026latent}, and InSight-o3~\cite{li2025insighto3}, further demonstrate the importance of iterative visual inspection, adaptive zooming, and tool-based local evidence acquisition. 
Related benchmarks and agents, such as AgentVista~\cite{su2026agentvista}, VTC-Bench~\cite{zhu2026vtcbench}, TIR-Bench~\cite{li2025tirbench}, and O3-Bench~\cite{li2025insighto3}, also highlight the need to evaluate long-horizon visual tool use and fine-grained multimodal reasoning. 
However, existing work still does not fully isolate the ability to repeatedly inspect fine-grained visual evidence, revisit image regions, and construct deep visual evidence chains.

\subsection{Benchmarking for Visual DeepSearch}

Existing benchmarks cover a range of tasks.
MMSearch~\cite{jiang2025mmsearch} and MMSearch-Plus~\cite{tao2026mmsearchplus} evaluate multimodal search and browsing abilities. MM-BrowseComp~\cite{li2025mmbrowsecomp}, VisBrowse-Bench~\cite{zhang2026visbrowse}, BrowseComp-$V^3$~\cite{zhang2026browsecompv3}, and VDR-Bench~\cite{zeng2026vdrbench} further emphasize multimodal browsing, visual-native search, and verifiable visual-textual evidence seeking. 
InterLV-Search~\cite{hou2026interlvsearch} studies interleaved language-vision agentic search, where visual evidence serves as an intermediate search pivot. 
MMDeepResearch-Bench~\cite{huang2026mmdeepresearch} focuses on multimodal deep research and citation-grounded report generation~\cite{ma2026tvir}. BEARCUBS~\cite{song2025bearcubs} evaluates live web-based information seeking with multimodal interactions.

Other benchmarks evaluate general capabilities. 
VTC-Bench~\cite{zhu2026vtcbench} focuses on compositional visual tool chaining, AgentVista~\cite{su2026agentvista} evaluates realistic multimodal agent interaction, ARK~\cite{lin2026ark} studies reasoning-aware multimodal retrieval~\cite{yang2026oacir}, and DeepWideSearch~\cite{lan2025deepwidesearch} analyzes depth-width trade-offs in agentic information seeking. 
Hierarchical lexical retrieval methods further support multi-hop evidence acquisition~\cite{ghassel2025hierarchical}.
Multi-hop and fine-grained visual reasoning benchmarks further examine structural-knowledge VQA~\cite{tran2025reasonvqa}, multi-entity multi-hop VQA~\cite{ma2026m3vqa}, ultra-high-resolution image reasoning~\cite{li2026urbench}, fine-grained visual observation~\cite{ye2025blinktwice,li2026fikabench,jiang2026pix2fact,li2026viebench}, visual multi-tabular reasoning~\cite{singh2025mtabvqa}, visual factuality evaluation~\cite{gu2025chinesesimplevqa}, and organic multimodal reasoning~\cite{hao2025emma}.

However, they remain insufficient for evaluating Visual DeepSearch capabilities. 
First, most of their tasks are not fully vision-centric: images often serve as initial context, while later reasoning can proceed through text, web evidence, or parametric knowledge. 
Second, fine-grained image retrieval and repeated region-level inspection are still under-evaluated. 
Third, many tasks have limited search horizons, making them solvable through shallow visual recognition or shortcut reasoning. 
Fourth, some datasets are vulnerable to knowledge leakage, where targets can be inferred from textual hints or memorized associations. 
Finally, transparent and reusable construction pipelines are still not consistently provided. 
To address these gaps, we introduce \bench, a benchmark designed to evaluate iterative, fine-grained, and evidence-grounded Visual DeepSearch.

\section{\bench Benchmark}
\sloppy

\begin{figure*}[t] 
    \centering
    \includegraphics[width=0.98\linewidth]{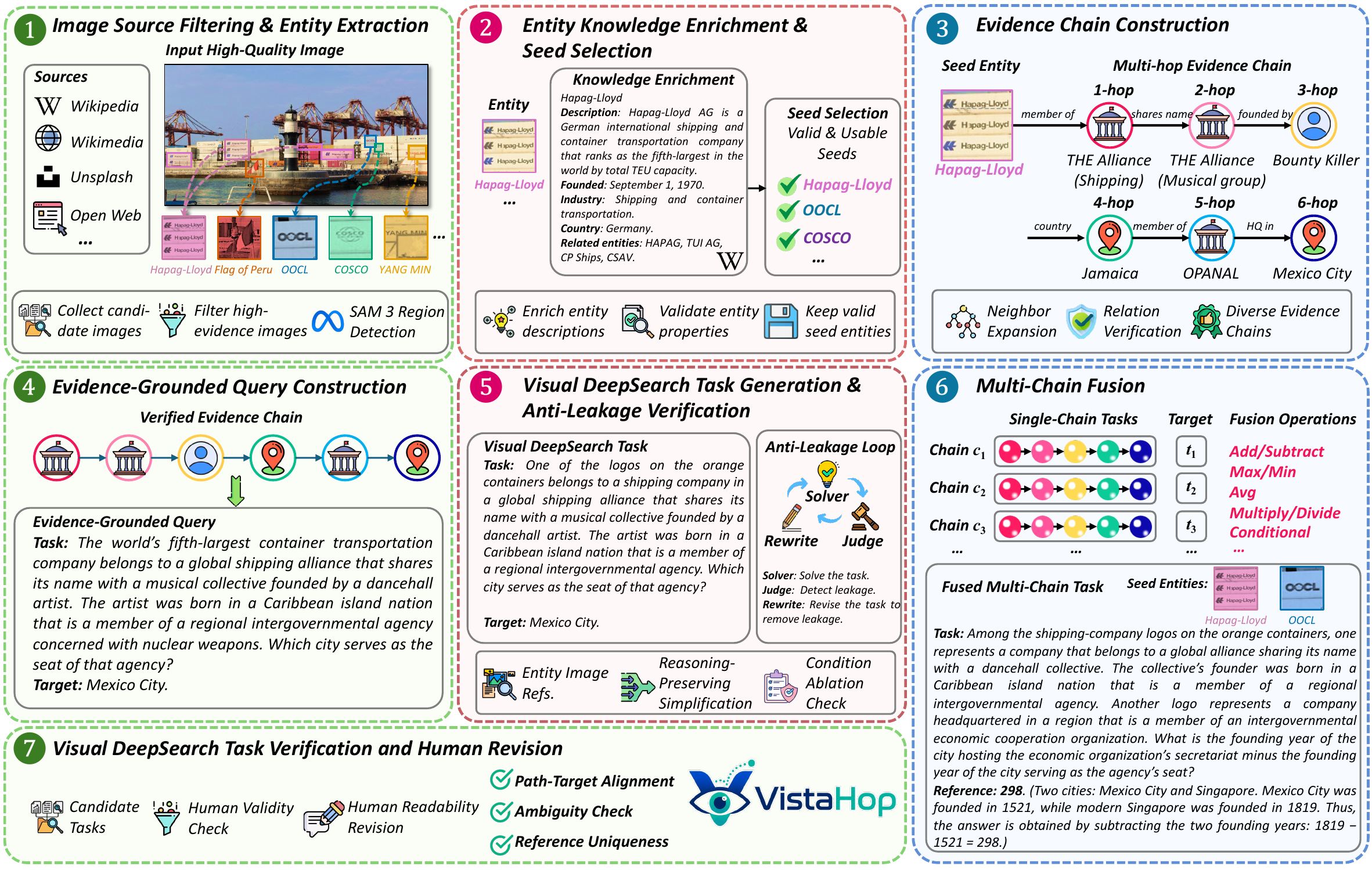}
    \caption{Overview of the \bench Construction Pipeline}
    \label{fig:seed}  
    % \vspace{-15pt}
\end{figure*}

\subsection{Benchmark Construction}
\label{sec:benchmark_construction}

As illustrated in Figure~\ref{fig:seed}, we construct \bench using a seven-stage pipeline.
\subsubsection{Image Source Filtering and Entity Extraction}
\label{sec:stage1}

\begin{figure}[t]
    \centering
    \includegraphics[width=\linewidth]{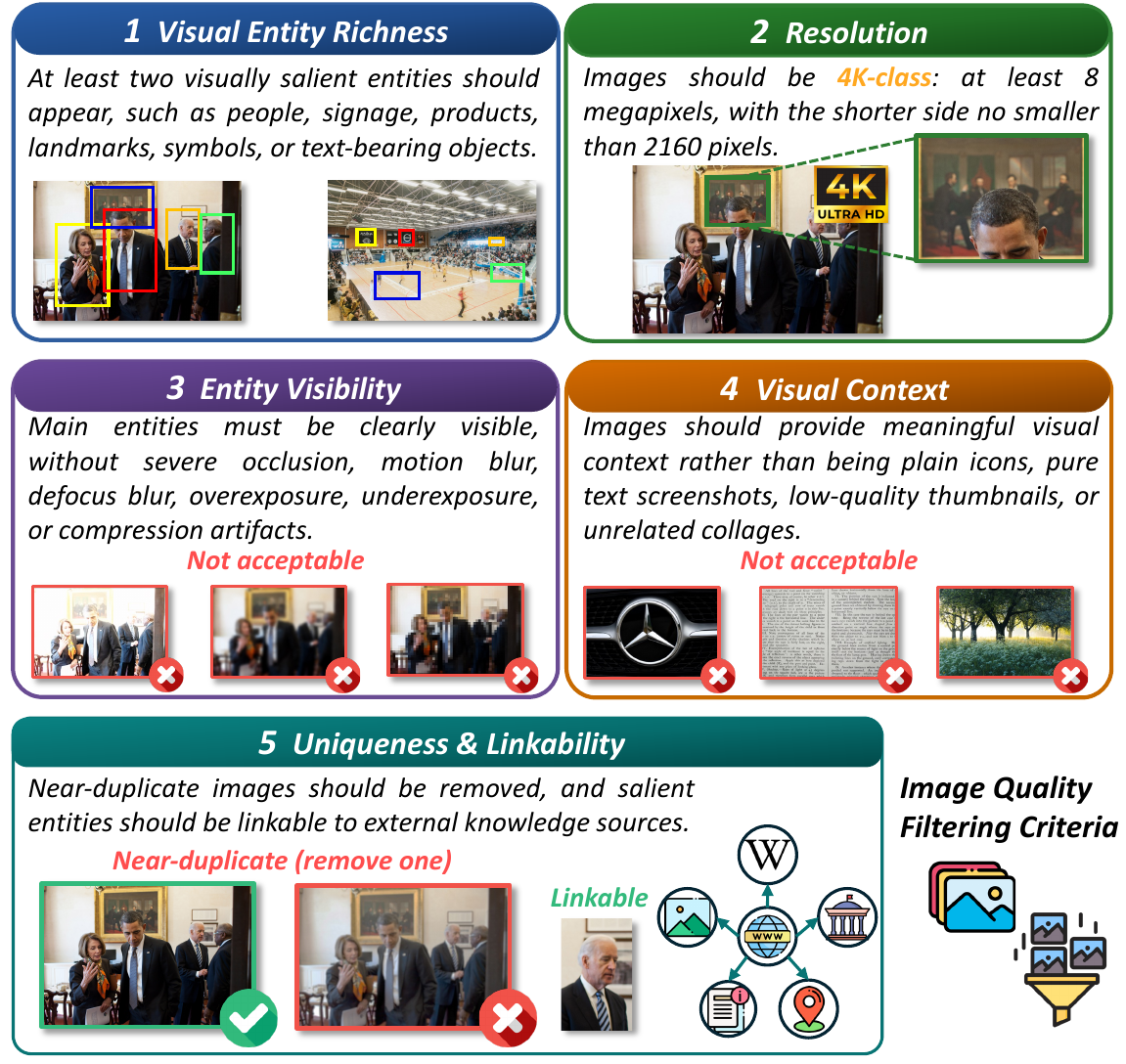}
    \caption{Image quality filtering criteria.}
    \label{image_filtering}
    % \vspace{-10pt}
\end{figure}

Candidate images are collected from publicly available and verifiable sources, including Wikimedia Commons~\cite{wikimedia2026commons}, Wikipedia pages~\cite{wikipedia2026}, Unsplash~\cite{unsplash2026}, and other open-access web sources with clear entity-level visual content.
To support reliable Visual DeepSearch, we retain only high-resolution images with sufficient visual evidence, recognizable entities, and local details for region-level inspection (details in Figure~\ref{image_filtering}).

We then extract image-grounded entities as entry points for downstream reasoning. 
Given an image, we first use \textsc{SAM~3}~\cite{carion2025sam3} to detect and segment candidate entity regions, obtaining an instance mask and bounding box for each region. We crop the detected regions and feed them to \textsc{Gemini-3.1-Pro}~\cite{google2026gemini31pro}, which performs OCR, entity naming, type classification, and visual attribute extraction. 
For each candidate region, the resulting structured record contains its mask and bounding box, a raw label, a canonical entity name, an entity type, a confidence score, and local visual attributes. 
The entity types include \textit{Person}, \textit{Organization}, \textit{Product}, \textit{Location}, \textit{Symbol}, and \textit{Event}, while the visual attributes capture appearance, color, scene context, logo patterns, and other discriminative cues.
We discuss construction-model sensitivity in Appendix~\ref{app:construction_model_sensitivity}.

We further normalize entity names, resolve aliases, remove duplicate detections, and filter low-confidence entities. 
Each retained entity is explicitly tied to a visual region and serves as a visual anchor for subsequent evidence-chain construction.

\subsubsection{Entity Knowledge Enrichment and Seed Selection}
\label{sec:stage2}

After extracting image-grounded entities, we enrich each entity with Wikipedia-derived textual knowledge and select valid seed entities for evidence-chain construction. 
For each entity, we query Wikipedia using its normalized name and retrieve its page content. 
An LLM then extracts a compact structured record, including a short description, an inferred entity type, useful properties, and related entities.
These enriched records form the initial seed pool. 
Entities with valid descriptions or properties are retained as seed candidates. 
If Wikipedia is insufficient, Wikidata-based properties are supplemented. 
Entities that still lack valid properties are discarded, as they cannot support reliable neighbor retrieval or evidence-chain expansion.

\subsubsection{Evidence Chain Construction}
\label{sec:stage3}

For each image-grounded seed, we construct a long-horizon evidence chain $c$ to a target, represented as
\begin{equation}
    c = (v_0 \xrightarrow{r_1} v_1 \xrightarrow{r_2} \cdots \xrightarrow{r_k} v_k),
\end{equation}
where $v_0$ and $v_k$ are the visual seed and target, and $r_i$ relates adjacent entities. Requiring at least five steps prevents shallow lookup.

\paragraph{Hidden reference subgraph.}
We view the open Web as an implicit multimodal evidence graph $\mathcal{G}_{\mathrm{web}}$, whose relations must be recovered from noisy webpages, images, and visually embedded text. Wikipedia and Wikidata are used only during construction to select a finite, verifiable reference subgraph $\mathcal{G}^{\mathrm{wiki}}_i\subseteq\mathcal{G}_{\mathrm{web}}$ and bound each task's answer. This subgraph and its annotated path are hidden from the solver. Given only the image, query, and search tools, the solver must recover a sufficient path step by step from unstructured textual and non-textual evidence. A different source-backed path remains valid if it uniquely supports the same target.

For each seed, we retrieve Wikidata neighbors and record every edge's entity identifiers, original predicate and direction, qualifiers, supporting evidence, and provenance. Predicates are normalized for analysis into seven categories: \textit{part-whole}, \textit{member-collection}, \textit{causal}, \textit{temporal}, \textit{spatial}, \textit{comparative}, and \textit{attributive}. We retain at most 30 neighbors, from which an LLM selects up to 10 context-relevant candidates.

Chains are expanded depth-first, with each edge validated against its retrieved evidence. We reject unsupported edges, repeated nodes, and chains with a shorter verified seed-to-target path in the collected evidence graph. Valid chains are ranked by entity-type and relation diversity:
\begin{equation}
    s_{\mathrm{div}}(c)
    ={} \alpha \frac{|\mathcal{T}_c|}{\min(k+1,|\mathcal{T}|)}
    + \beta \frac{|\mathcal{R}_c|}{\min(k,7)},
\end{equation}
where $\mathcal{T}_c$ and $\mathcal{R}_c$ are the entity and relation types in $c$, and $\mathcal{T}$ is the Stage~3 entity-type inventory. Both terms are normalized to $[0,1]$. We set $\alpha=0.6$ and $\beta=0.4$ and retain the chains with the highest $s_{\mathrm{div}}$ scores.

Finally, an LLM merges adjacent aliases or spelling variants. After each merge, we re-verify affected edges, recompute length and diversity, and reapply the $k\geq5$ and no-shortcut constraints. We denote the resulting set of retained evidence chains by $\mathcal{C}$:
\begin{equation*}
    \mathcal{C} = \{c_1, c_2, \ldots, c_m\}.
\end{equation*}
Each chain $c_j$ records its seed, intermediate nodes, original and normalized relations, directions, qualifiers, evidence and provenance, verification confidence, familiarity scores, and its step count $k_j$.

\subsubsection{Evidence-Grounded Query Construction}
\label{sec:stage4}

Given a verified evidence chain 
$c_j$, 
we generate a natural-language query $q_j$ about its terminal node $v_{j,k_j}$ and use the node's canonical name or normalized value as the target.

The query is written in an indirect form to reduce direct string matching, entity lookup, and text-only shortcuts. 
We describe the seed or intermediate entities using visual or knowledge-grounded clues, such as ``\textit{the organization whose logo appears in the image}.'' 
The output of this step is a set of candidate evidence-grounded query instances:
\begin{equation*}
    \mathcal{Q}
    =
    \{(q_j, t_j, c_j)\}_{j=1}^{m},
\end{equation*}
where $q_j$ is the generated query, $t_j$ is the target, and $c_j$ is the corresponding evidence chain.

\subsubsection{Visual DeepSearch Task Generation and Anti-Leakage Verification}
\label{sec:stage5}

For each candidate $(q_j,t_j,c_j)\in\mathcal{Q}$, we attach image $I_j$, convert $c_j$ into a structured reasoning path $\rho_j$, and simplify the query to $\tilde{q}_j$ while preserving its target and reasoning chain.

\paragraph{Multi-agent anti-leakage verification.}
We use a three-agent loop to detect textual shortcuts. Given only $\tilde{q}_j$, the \textsc{Solver} attempts to recover the target without the image or chain. The \textsc{Judge} assesses correctness, leakage, ambiguity, under-specification, and uniqueness. If needed, the \textsc{Rewrite} agent revises the query while preserving its image, target, and chain. Unresolved queries are discarded after at most $R_{\max}=3$ rounds.

\paragraph{Clue-necessity verification.}
Separately, we test whether each atomic evidence-bearing clue is necessary. We map clues to edges or constraints, ablate them individually, and recompute target reachability and uniqueness in the evidence graph. A blind image-and-search-enabled \textsc{Solver} also evaluates the original and ablated queries under the same tool budget, without access to the target or annotated chain. Solver evidence is considered only when the same Solver recovers the original target; failure on an ablation alone does not establish necessity. A clue is redundant if the remaining constraints uniquely identify the target, enable a verified alternative path, or still allow paired Solver recovery. It is necessary only if ablation removes unique graph reachability and yields no paired recovery. We manually review ambiguous cases, remove redundant clues, recompute the path and step count, and rerun anti-leakage verification. Invalid items are rewritten or discarded.

The final task item is represented as:
\begin{equation*}
    x_j = (I_j, \tilde{q}_j, t_j, \rho_j, M_j),
\end{equation*}
where $M_j$ stores its step count, relation types, source entity, and verification and generation records. The retained items form $\mathcal{D}_{\text{single}}=\{x_j\}_{j=1}^{n_{\text{single}}}$, where $n_{\text{single}}\leq m$. Each requires its intended image-grounded chain and is not reliably solvable from text alone.

\subsubsection{Multi-Chain Fusion}
\label{sec:stage6}

To broaden the search space, we fuse $k$ ($k\geq2$) thematically or logically related tasks so that the model must start from different image-grounded points, recover all component results, preserve the correct associations between anchors and values, and compose them correctly. From each component target or metadata, we extract and normalize an intermediate value and unit (\eg, year, count, ranking, or quantity). A deterministic program applies a semantically appropriate operation, such as addition, subtraction, maximum, average, or conditional selection. The operation is intentionally simple because the purpose is to broaden the model's search rather than test arithmetic complexity. We reject ambiguous values, incompatible units, undefined orderings, and invalid arithmetic. The final query retains each component's visual or evidence-grounded clue while hiding its intermediate value.

Post-fusion verification reruns the text-only shortcut check, confirms unique normalized component targets and a single-valued result, and applies clue-necessity testing by ablating each complete component clue. Every component must also have a verified visual anchor. For each anchor--component pair, the fixed visual-dependence scorer compares the component target under the original image and an image with that anchor masked. The 0.182 log-likelihood-gap threshold in Section~\ref{sec:benchmark_quality} is an operational screen, not a causal criterion; pairs at or below it undergo adjudicated review to confirm that masking removed the seed evidence. Items with missing coverage or failed review are rewritten and fully reverified or discarded. Retained items store the query, deterministic target, extraction and calculation records, components, necessity and masking records, and fused path. They form the fused-task set $\mathcal{D}_{\text{fused}}$. The final benchmark dataset is $\mathcal{D}=\mathcal{D}_{\text{single}}\cup\mathcal{D}_{\text{fused}}$.

\subsubsection{Visual DeepSearch Task Verification and Human Revision}
\label{sec:stage7}
As a final quality-control step, three human annotators manually verify each candidate. 
They check whether the reasoning path leads to the annotated target, whether the query is ambiguous or misleading, and whether the target is unique. Instances that fail any check are discarded.
The annotators then revise the retained task queries for readability while preserving the visual grounding, reasoning path, and final target. All automatic validity, clue-necessity, anti-leakage, and visual-dependence checks are rerun after revision; instances that fail a hard constraint are discarded. Table~\ref{tab:chain_audit} in Appendix~\ref{app:qc_filtering_composition} reports pre-revision structural checks and post-revision query naturalness.

% \paragraph{Dependence on construction models.}
% Construction models propose and filter candidate regions, chains, and queries, so replacing them may change the benchmark's exact composition. Their outputs are not accepted directly as ground truth: every retained item must pass the same source-support, no-shortcut, target-uniqueness, anti-leakage, clue-necessity, visual-dependence, and human-review requirements, together with deterministic calculation where applicable. Model choice therefore affects which candidate subgraphs are sampled, while the fixed acceptance protocol controls their validity.

\subsection{Benchmark Quality}
\label{sec:benchmark_quality}
\paragraph{Visual dependence.}
We measure whether a task requires visual evidence using the target sequence log-likelihood gap between the original and heavily masked images:
\begin{equation}
    s_{\mathrm{vis}}^{(j)}=
    \log \pi_{\theta}(t_j \mid \tilde{q}_j,I_j)
    -
    \log \pi_{\theta}(t_j \mid \tilde{q}_j,I_{j,\mathrm{mask}}) .
\end{equation}
Here, $\pi_{\theta}$ is the fixed Qwen3-VL-32B-Instruct scorer~\cite{bai2025qwen3vl}, $t_j$ is the target sequence, and $\tilde{q}_j$ is the final query. A larger gap indicates stronger visual dependence. We manually inspect items with $s_{\mathrm{vis}}^{(j)} \leq 0.182$ and rewrite or discard those solvable from textual clues alone. Scores are recomputed after all revisions. The final set has a median score of 0.770, with 92.3\% of items above 0.182.
Appendix~\ref{app:visual_dependence_threshold} explains the selected threshold,
its likelihood-ratio interpretation, and its model-specific scope.

\paragraph{Task difficulty.}
We define task difficulty by the evidence-step count $H$: the number of relational transitions to the target, summed across component chains for fused tasks.

\begin{table}[t]
\centering
\scriptsize
\caption{Difficulty levels in \bench.}
\label{tab:difficulty_levels}
\renewcommand{\arraystretch}{0.8}
\setlength{\tabcolsep}{4pt}
\begin{tabularx}{\linewidth}{p{0.09\linewidth} p{0.15\linewidth} X}
\toprule[1.5pt]
% \rowcolor[gray]{0.8}
\textbf{Level} & \textbf{Evidence Steps} & \textbf{Description} \\
\midrule
L1
& $H\in[1,5)$
& Basic tasks that require fewer than five evidence steps. These tasks mainly evaluate visual grounding and short-range evidence connection. \\
\midrule
L2
& $H\in[5,10)$
& Medium-difficulty tasks that require long-chain evidence traversal. These tasks evaluate whether models can track multiple intermediate entities and relations. \\
\midrule
L3
& $H\in[10,\infty)$
& Hard tasks that require very long or multiple evidence paths. These tasks evaluate complex evidence coordination, long-horizon evidence traversal, and compositional target derivation. \\
\bottomrule[1.5pt]
\end{tabularx}
% \vspace{-8pt}
\end{table}

% \begin{figure}[t]
% \centering
% \includegraphics[width=0.8\linewidth]{figure/difficulty-levels.pdf}
% \caption{Difficulty levels in \bench.}
% \label{fig:difficulty_levels}
% \vspace{-15pt}
% \end{figure}

\subsection{Benchmark Statistics}
\label{sec:benchmark_statistics}
\begin{figure}[t]
    \centering
    \includegraphics[width=0.8\linewidth]{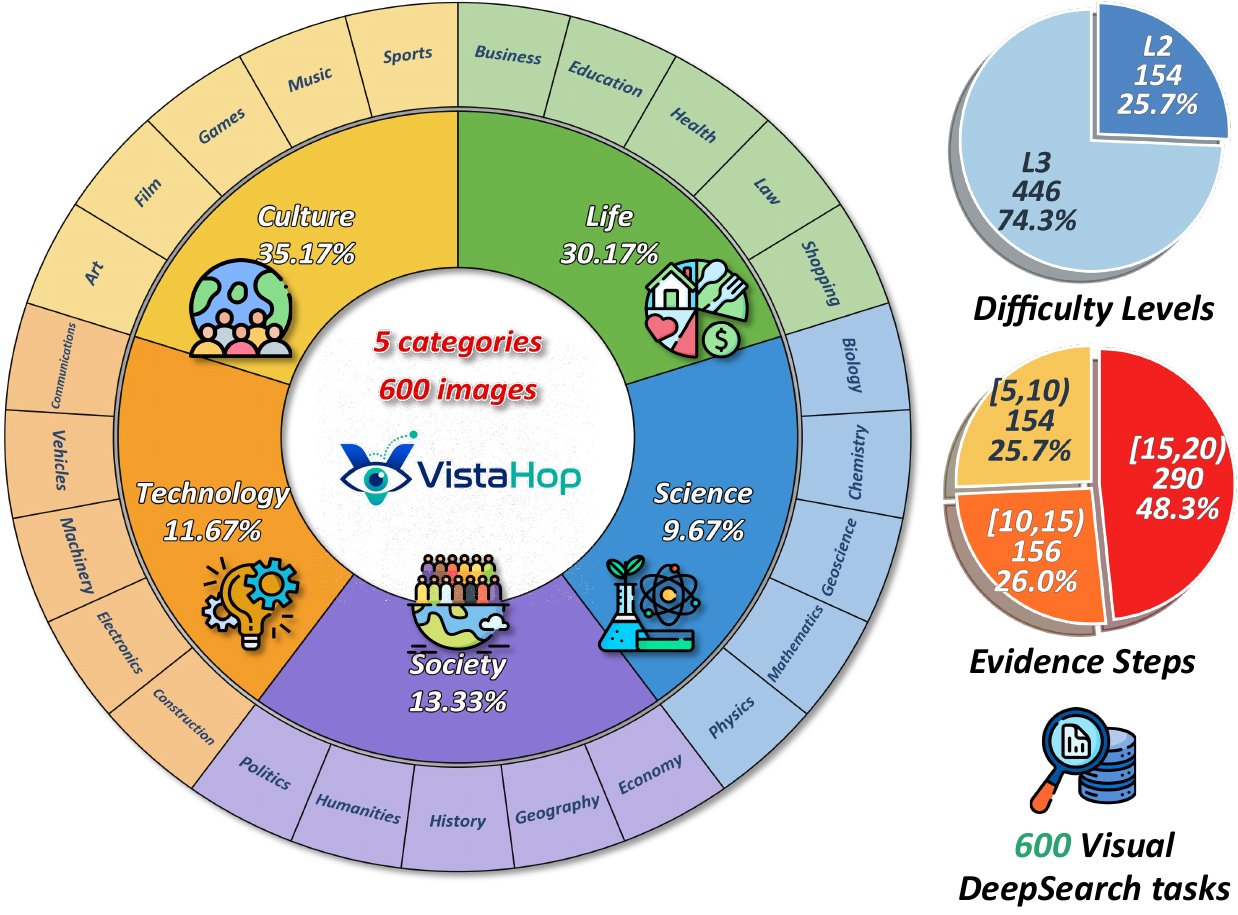}
    \caption{Overall statistics of \bench.}
    \label{fig:statistics}
    % \vspace{-8pt}
\end{figure}

We categorize each image according to the dominant visual content and real-world context depicted in the image. \bench contains 600 high-resolution images covering 25 visual search scenarios in 5 categories: \textit{Life}, \textit{Science}, \textit{Society}, \textit{Technology}, and \textit{Culture}. After entity extraction, we obtain 5,184 image-grounded entities and retain 3,348 seed entities. Across the retained tasks, the annotations contain 1,330 component evidence chains. Aggregating component-chain steps at the question level gives an average of 14.92 evidence steps per task, with every task requiring at least 5 steps. 

\begin{figure}[t]
    \centering
    \includegraphics[width=\linewidth]{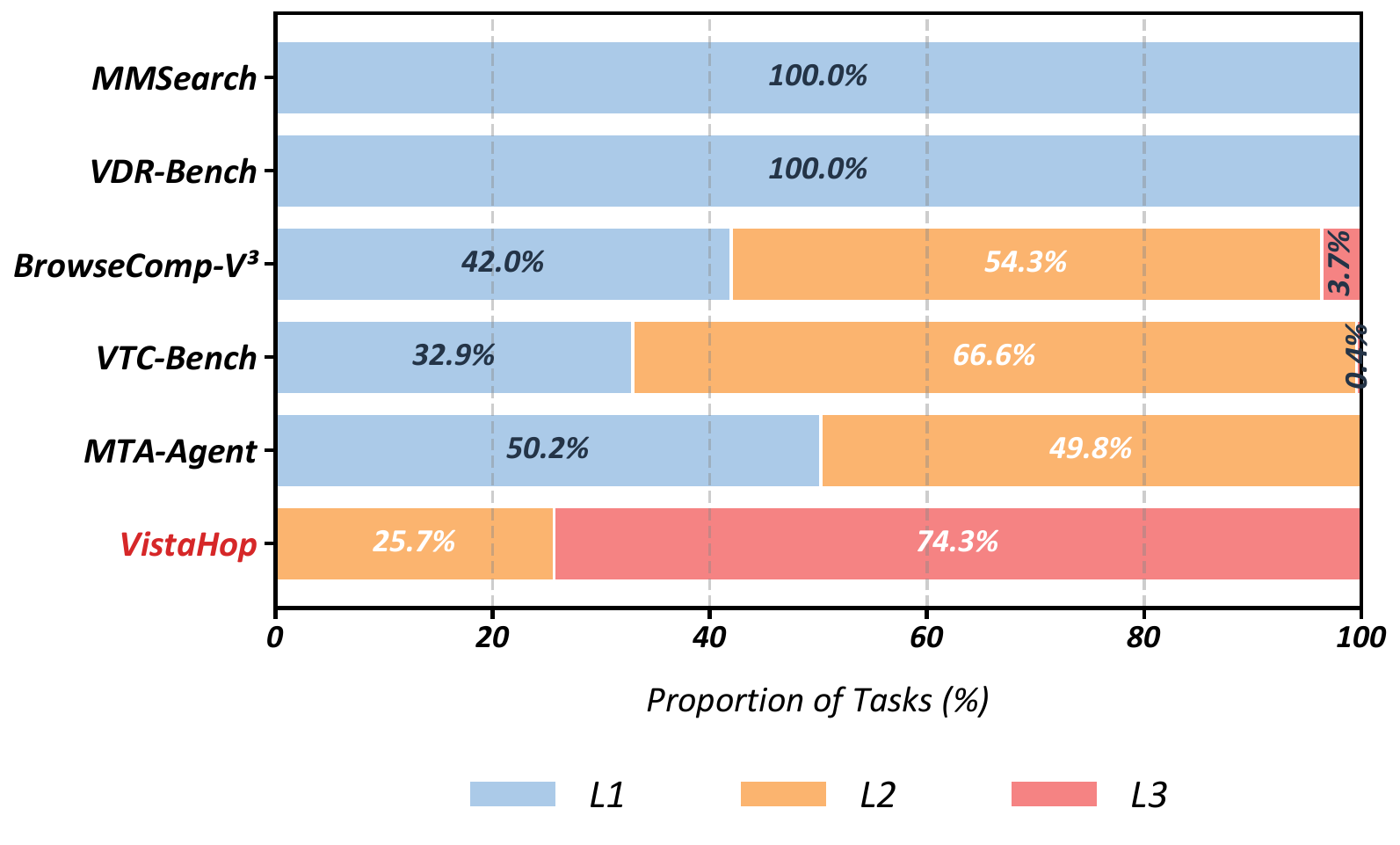}
    \caption{Difficulty-level distributions across benchmarks.}
    \label{fig:baseline_difficulty_distribution}
    % \vspace{-12pt}
\end{figure}

Figure~\ref{fig:baseline_difficulty_distribution} compares \bench with
MMSearch~\cite{jiang2025mmsearch}, VDR-Bench~\cite{zeng2026vdrbench},
BrowseComp-$V^3$~\cite{zhang2026browsecompv3}, VTC-Bench~\cite{zhu2026vtcbench},
and MTA-Agent~\cite{peng2026mtaagent}. We normalize
their hop counts using the cross-benchmark annotation protocol detailed in
Appendix~\ref{appendix:cross_benchmark_hops}. These benchmark datasets are
heavily concentrated at the L1 and L2 levels: L3 instances account for only
3.7\% of BrowseComp-$V^3$ and 0.4\% of VTC-Bench. By contrast, \bench consists
of L2 and L3 tasks, with 74.3\% categorized as L3, thereby placing substantially
greater emphasis on long and compositional evidence chains.

\section{\eval Evaluation Framework}
\begin{figure*}[t]
    \centering
    \includegraphics[width=\linewidth]{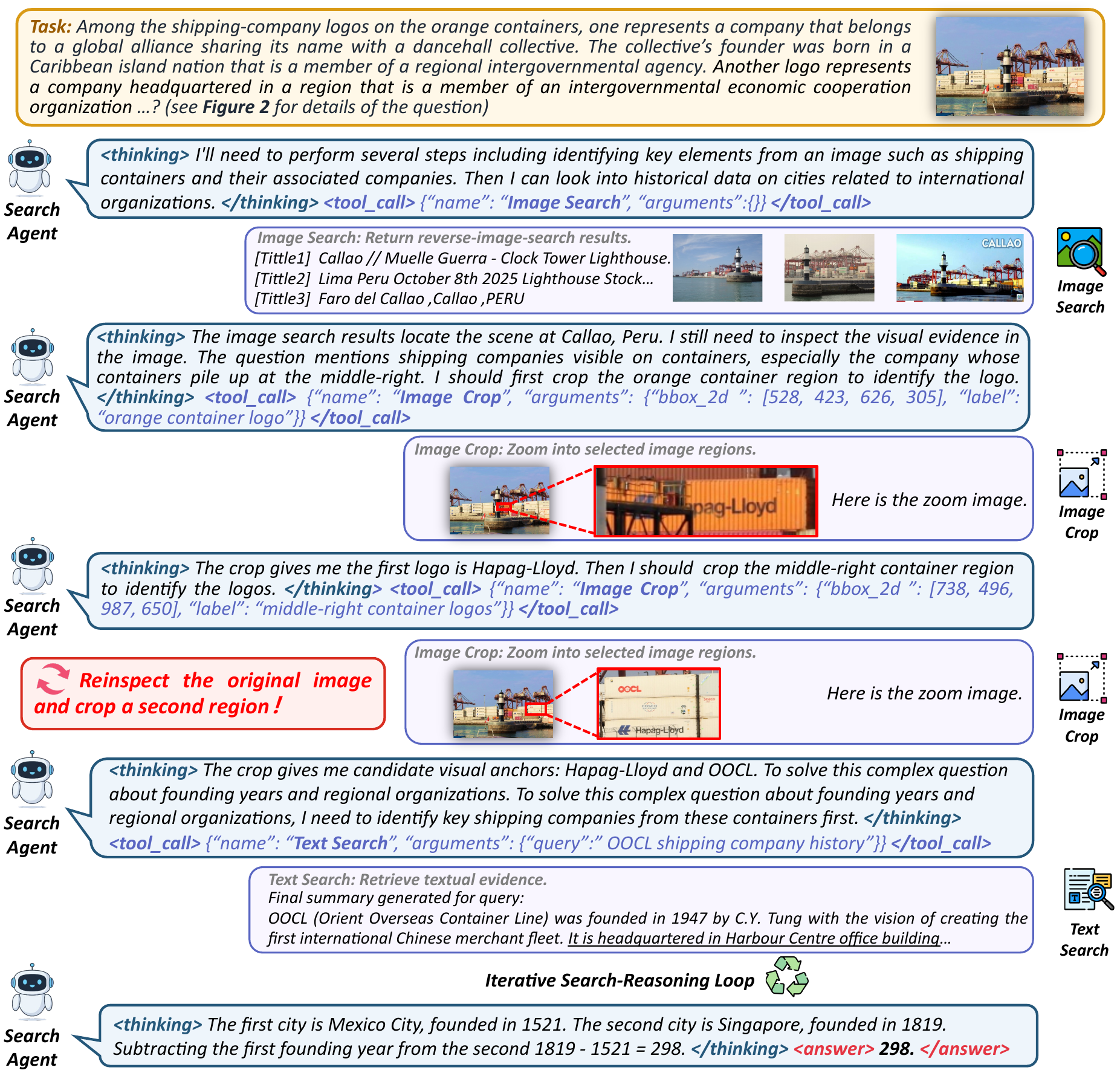}
    \caption{An illustrative Visual DeepSearch trajectory in \eval. See Appendix~\ref{appendix:expanded_search_example} for the full loop.}
    \label{fig:evaluation_trajectory}
    \vspace{-10pt}
\end{figure*}

\begin{figure}[t]
    \centering
    \includegraphics[width=\linewidth]{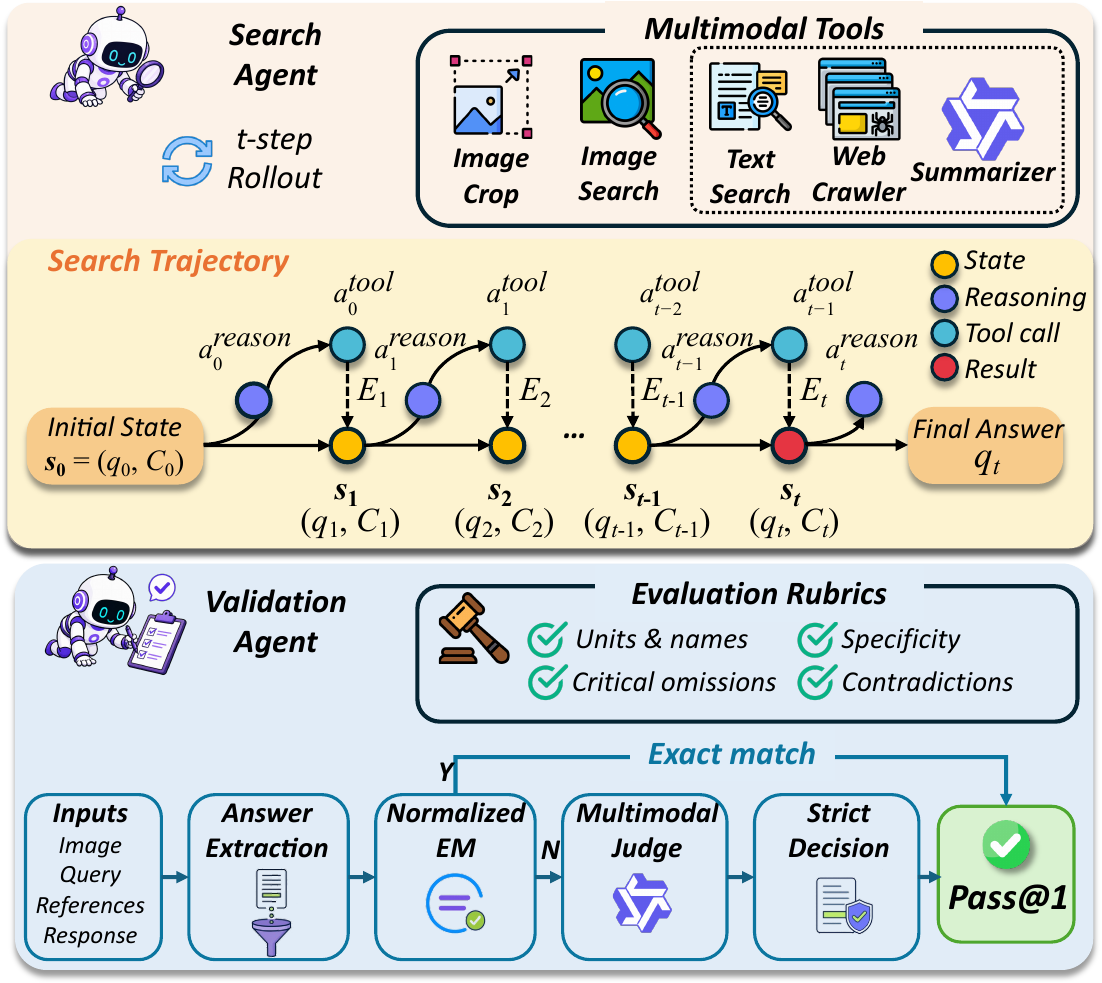}
    \caption{Overview of \eval. Search Agent performs iterative visual inspection, retrieval, and evidence-grounded reasoning; Validation Agent evaluates the resulting trajectory and final answer.}
    \label{fig:evaluation_overview}
    % \vspace{-10pt}
\end{figure}

We design \eval for automated testing on \bench. It contains a \emph{Search Agent} for tool-augmented reasoning and a \emph{Validation Agent} for response checking (Figures~\ref{fig:evaluation_trajectory} and~\ref{fig:evaluation_overview}).

\textbf{Search Agent.} 
The agent iteratively observes the query, image, history, and accumulated evidence, then either invokes one tool or submits a response. It can use \textit{Text Search} for external facts, \textit{Image Search} for reverse-image evidence, and \textit{Image Crop} for local inspection. \textit{Text Search} crawls detailed webpages and summarizes their content. The reference graph, annotated path, and target are hidden; tool results instead contain noisy, unstructured textual and visual evidence. The agent must extract the relevant entities and relations and assemble a sufficient path step by step. The process runs for at most $N$ rounds, with at most one tool call per round. Figure~\ref{fig:evaluation_trajectory} abbreviates this iterative search--reasoning loop.

\medskip
\textbf{Validation Agent.}
Given the image, query, reference answers, and first submitted response, the Validation Agent extracts the final answer and first applies normalized exact matching. Exact matches pass directly; otherwise, the inputs are sent to the multimodal judge for a strict binary decision that accepts only semantically correct, complete, and unambiguous answers. \emph{Pass@1} records whether the first response is accepted; we also report average tool calls, average rounds, and tool utilization, defined as average tool calls divided by the maximum allowed calls.

Each instance is evaluated five times and averaged; systems are anonymized, query order is randomized, and the judge model is distinct from all evaluated models~\cite{liu2025elo,liufewer}.

\begin{table}[t]
\centering
\tiny
\newcommand{\modelicon}[2][1.25em]{\raisebox{-0.2ex}{\includegraphics[height=#1]{#2}}\hspace{0.3em}}
\caption{Ablation study of tool integration on \bench. 
P@1 denotes Pass@1. \emph{No-Tool} uses no external tools, \emph{Search} enables \textit{Text Search} and \textit{Image Search}, and \emph{Search+Crop} further enables image cropping.
}
\label{tab:model_performance}
\setlength{\tabcolsep}{1.6pt}
\renewcommand{\arraystretch}{0.9}
\resizebox{\linewidth}{!}{
\begin{tabular}{@{}ll|ccc|ccc@{}}
\toprule[1.2pt]
\multirow{2}{*}{\textbf{Model}} 
& \multirow{2}{*}{\textbf{Set.}}
& \multicolumn{3}{c|}{\textbf{Response}}
& \multicolumn{3}{c}{\textbf{Process}} \\
\cline{3-8}
& 
& \textbf{P@1}
& \textbf{L2}
& \textbf{L3}
& \textbf{Calls}
& \textbf{Rnds}
& \textbf{Util.}
\\
\specialrule{\lightrulewidth}{\aboverulesep}{0pt}
\rowcolor{gray!30}
\multicolumn{8}{c}{\textit{Closed-Source Models}} \\
\specialrule{\lightrulewidth}{0pt}{0pt}

\multirow{3}{*}{\modelicon{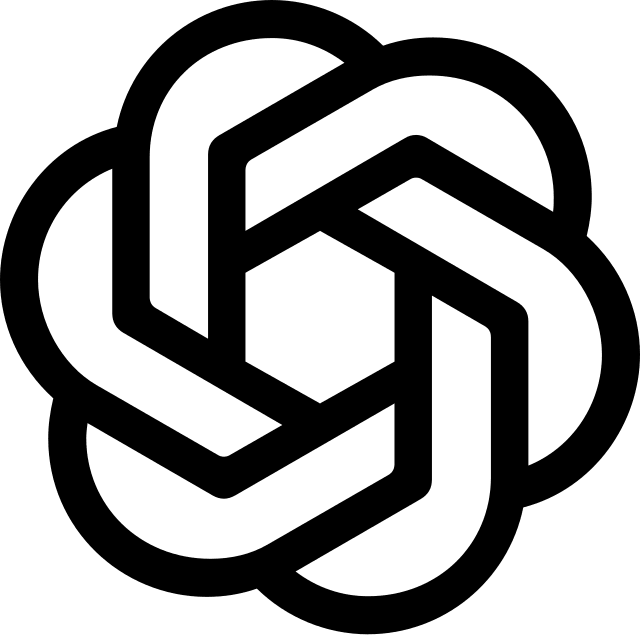}GPT-5.2}
& \emph{No-Tool}     & 11.33 & 27.92 & 5.61 & 0.00 & 1.00 & 0.0 \\
& \emph{Search}      & 15.83 & 37.66 & 8.30 & 7.88 & 8.72 & 78.8 \\
& \emph{Search+Crop} & 25.83 & \cellcolor{red!25}\textbf{40.91} & 20.63 & 8.66 & 9.16 & 86.6 \\
\midrule

\multirow{3}{*}{\modelicon{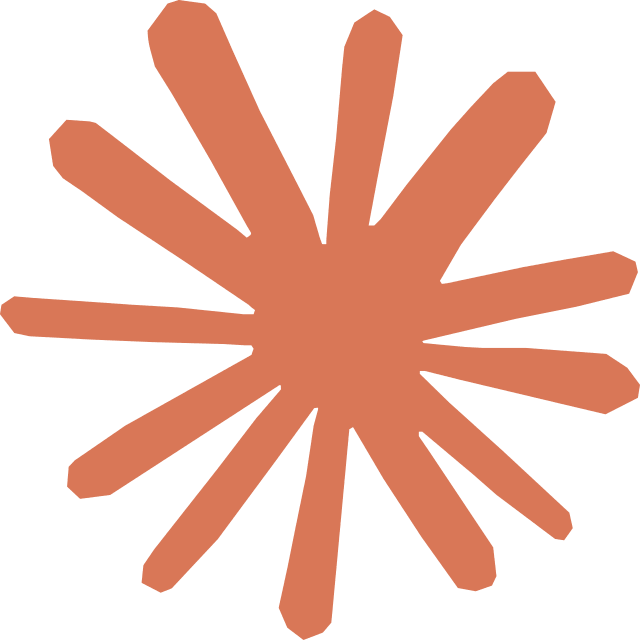}\shortstack[l]{Claude Sonnet\\4.5}}
& \emph{No-Tool}     & 11.08 & 27.27 & 5.49 & 0.00 & 1.00 & 0.0 \\
& \emph{Search}      & 14.47 & 32.29 & 8.32 & 7.63 & 8.49 & 76.3 \\
& \emph{Search+Crop} & 23.68 & 37.03 & 19.07 & 8.39 & 9.23 & 83.9 \\
\midrule

\multirow{3}{*}{\modelicon{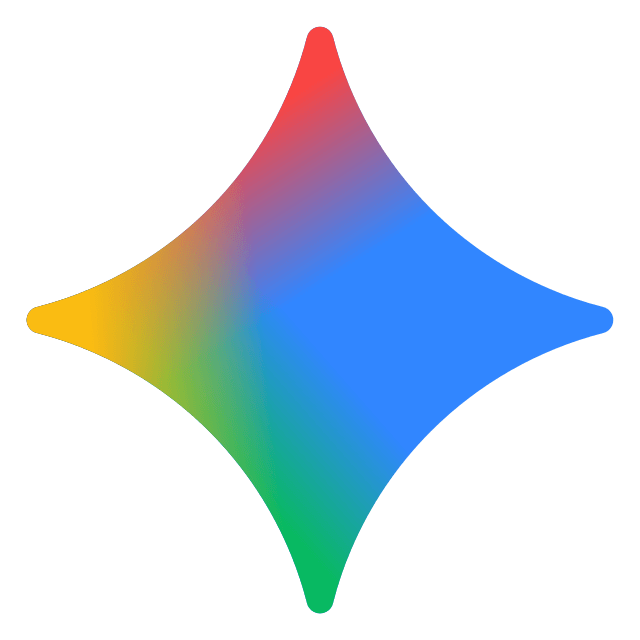}Gemini 2.5 Pro}
& \emph{No-Tool}     & 8.63 & 21.33 & 4.24 & 0.00 & 1.00 & 0.0 \\
& \emph{Search}      & 13.74 & 29.97 & 8.14 & 6.89 & 7.81 & 68.9 \\
& \emph{Search+Crop} & 21.83 & 34.15 & 17.58 & 7.78 & 8.67 & 77.8 \\
\specialrule{\lightrulewidth}{\aboverulesep}{0pt}
\rowcolor{gray!30}
\multicolumn{8}{c}{\textit{Open-Source Models}} \\
\specialrule{\lightrulewidth}{0pt}{0pt}

\multirow{3}{*}{\modelicon{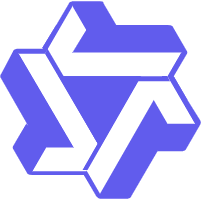}Qwen3-VL-235B}
& \emph{No-Tool}     & 7.84 & 19.04 & 3.97 & 0.00 & 1.00 & 0.0 \\
& \emph{Search}      & 13.13 & 28.06 & 7.97 & 7.26 & 8.21 & 72.6 \\
& \emph{Search+Crop} & 20.54 & 32.09 & 16.55 & 8.08 & 8.97 & 80.8 \\
\midrule

\multirow{3}{*}{\modelicon{figure/logos/qwen.png}Qwen3-VL-30B}
& \emph{No-Tool}     & 6.72 & 15.69 & 3.62 & 0.00 & 1.00 & 0.0 \\
& \emph{Search}      & 11.63 & 24.91 & 7.04 & 6.44 & 7.41 & 64.4 \\
& \emph{Search+Crop} & 18.34 & 28.69 & 14.77 & 7.19 & 8.13 & 71.9 \\
\specialrule{\lightrulewidth}{\aboverulesep}{0pt}
\rowcolor{gray!30}
\multicolumn{8}{c}{\textit{Agentic Models}} \\
\specialrule{\lightrulewidth}{0pt}{0pt}

\multirow{3}{*}{\modelicon[0.85em]{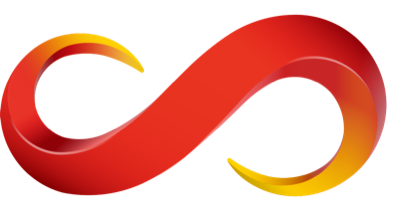}\shortstack[l]{SenseNova-\\MARS-32B}}
& \emph{No-Tool}     & 10.00 & 24.68 & 4.93 & 0.00 & 1.00 & 0.0 \\
& \emph{Search}      & 14.83 & 34.42 & 8.07 & 8.04 & 8.95 & 80.4 \\
& \emph{Search+Crop} & \cellcolor{red!25}\textbf{26.33} & 40.26 & \cellcolor{red!25}\textbf{21.52} & \cellcolor{red!25}\textbf{8.83} & \cellcolor{red!25}\textbf{9.45} & \cellcolor{red!25}\textbf{88.3} \\
\midrule

\multirow{3}{*}{\modelicon{figure/logos/qwen.png}\shortstack[l]{MMSearch-R1\\7B}}
& \emph{No-Tool}     & 5.78 & 13.27 & 3.19 & 0.00 & 1.00 & 0.0 \\
& \emph{Search}      & 11.37 & 24.43 & 6.86 & 6.07 & 6.93 & 60.7 \\
& \emph{Search+Crop} & 17.18 & 27.46 & 13.63 & 6.81 & 7.57 & 68.1 \\
\midrule

% Tongyi-DeepResearch-30B-A3B is text-only and cannot consume the image input
% required by VistaHop, so its results are excluded from the multimodal comparison.
% \multirow{3}{*}{\modelicon{figure/logos/tongyi_deepresearch.png}Tongyi-DR$^\dagger$}
% & \emph{No-Tool}     & 5.40 & 12.83 & 2.83 & 0.00 & 1.00 & 0.0 \\
% & \emph{Search}      & 10.80 & 22.69 & 3.69 & 7.92 & 8.84 & 79.2 \\
% & \emph{Search+Crop} & 16.90 & 25.82 & 3.82 & 8.15 & 9.08 & 81.5 \\
% \midrule
Human
& -- & 79.50 & 81.17 & 78.92 & -- & -- & -- \\

\bottomrule[1.2pt]
\end{tabular}
}
% \vspace{-15pt}
\end{table}

\section{Experiments}
\subsection{Experimental Setup}
\label{sec:experimental_setup}
\textbf{Evaluated models.}
We evaluate 7 representative multimodal models for Visual DeepSearch: 
SenseNova-MARS-32B~\cite{chng2025sensenovamars},
GPT-5.2~\cite{openai2025gpt52}, 
Claude Sonnet 4.5~\cite{anthropic2025claudesonnet45},
Gemini 2.5 Pro~\cite{comanici2025gemini},
Qwen3-VL-235B-A22B-Instruct~\cite{bai2025qwen3vl,qwen2025qwen3vl235ba22binstruct},
Qwen3-VL-30B-A3B-Instruct~\cite{bai2025qwen3vl,qwen2025qwen3vl30ba3binstruct},
and MMSearch-R1-7B~\cite{wu2025mmsearchr1}.
SenseNova-MARS-32B and MMSearch-R1-7B are post-trained with agentic reinforcement learning.
% Tongyi-DeepResearch-30B-A3B is excluded because it accepts text only and
% therefore cannot directly process VistaHop's image-conditioned inputs.

\textbf{Evaluation models.}
We use Qwen3-VL-32B-Instruct~\cite{bai2025qwen3vl} as the judge model for automatic response assessment. 
% It is served by SGLang with bfloat16 precision and a context length of 40,960. 
We use Qwen3-32B~\cite{yang2025qwen3} as the summarizer.

\textbf{Settings.} Unless otherwise specified, \eval uses a maximum of $N=10$ reasoning rounds and at most one tool call per round. For \textit{Text Search}, the retrieval depth is fixed to the top-3 results; \textit{Image Search} returns up to 5 reverse-image-search results; and the default generation temperature is 0.7. All experiments are run on Intel Xeon Gold 5218 CPUs and $16\times$ H200-141G GPUs. Appendix~\ref{appendix:retrieval_reproducibility} documents the retrieval backends, caching policy, temporal scope, and handling of unavailable webpages.

\textbf{Human baseline.} We recruit three graduate students with computer science backgrounds and experience in multimodal reasoning and web search from the authors' institutions. None of them participated in benchmark construction. 
Each participant independently completes all 600 tasks using only the task query and image, with access to web search, image search, and image cropping but not the reference target or annotated evidence chain. Responses are evaluated using the same Pass@1 criterion, and we report the mean Pass@1 across the three participants. Appendix~\ref{appendix:human_baseline_protocol} provides the detailed protocol, difficulty-subset sizes, and aggregation procedure.
Appendix~\ref{appendix:ethics_licensing_responsible_use} discusses image
licensing, privacy, bias, and responsible-use considerations.

\subsection{Overall Performance}
As shown in Table~\ref{tab:model_performance}, the best model reaches only 26.33\% Pass@1, indicating that Visual DeepSearch remains difficult. Under \emph{Search+Crop}, SenseNova-MARS-32B achieves 40.26\% on L2 tasks but only 21.52\% on L3 tasks. Tool use is essential: average Pass@1 rises from 8.77\% in \emph{No-Tool} to 21.96\% in \emph{Search+Crop}. Under \emph{Search+Crop}, agents make 7.96 tool calls on average, while the annotated reasoning paths contain an average of 14.92 evidence steps per question, reflecting the substantial interaction demands of the benchmark.

Performance also drops clearly from L2 to L3: under \emph{Search+Crop}, average Pass@1 decreases from 34.37\% to 17.68\%. This gap indicates that tasks with longer and more compositional evidence chains place substantially greater demands on visual grounding, evidence tracking, and long-horizon search. SenseNova-MARS-32B obtains the highest accuracy, while MMSearch-R1-7B reaches 17.18\% with 6.81 tool calls on average, reflecting a more selective but less effective search policy on long-horizon, crop-intensive tasks. Nevertheless, both agentically post-trained models remain limited in visual grounding, evidence revisiting, and cross-chain reasoning.

\textbf{Human validation of automatic judging.} For 300 outputs spanning models, inference settings, and difficulty levels, three annotators' labels agree with the LLM-as-judge results in 90.8\% of cases, with strong inter-annotator agreement (Fleiss' $\kappa=0.88$).

\subsection{Ablation Study}
We compare \emph{No-Tool}, \emph{Search}, and \emph{Search+Crop} in
Table~\ref{tab:model_performance}. Search raises average Pass@1 from 8.77\% to
13.57\% ($+4.80$ points), while cropping further improves it to 21.96\%
($+8.39$ points). All seven models benefit from cropping, and
SenseNova-MARS-32B performs best at 26.33\%.
The larger crop gain indicates that fine-grained visual grounding and
evidence verification remain major bottlenecks beyond external retrieval.
Appendix~\ref{appendix:repeated_inspection_protocol} details how we measure
visual grounding and repeated image inspection and describes the one-shot,
dynamic-crop, masking, and distractor controls.
Appendix~\ref{appendix:nonindexed_images} tests robustness on previously
unpublished images that are not indexed by reverse-image search.

\subsection{Sensitivity Study}

Table~\ref{tab:sensitivity} studies the effect of the maximum number of reasoning rounds $N$. Because $H$ counts semantic evidence transitions rather than interaction rounds, it does not map one-to-one to $N$: agents may resolve some transitions using parametric knowledge and obtain multiple pieces of evidence from a single retrieval. Increasing $N$ improves performance up to $N=10$, where the model has enough budget to inspect visual anchors, search intermediate evidence, and verify its answer. Larger budgets introduce more redundant calls and noisy context, causing performance to saturate or decline~\cite{huang2026does,yue2025promoting}. This suggests that Visual DeepSearch requires not only more evidence, but also accurate early visual grounding and efficient search control. 

\begin{table}[!t]
\centering
\scriptsize
\caption{Sensitivity analysis of the maximum reasoning rounds $N$ on \bench using SenseNova-MARS-32B under the \emph{Search+Crop} setting.
% We report Pass@1, L2/L3 performance, and process-level statistics.
}
\label{tab:sensitivity}
\setlength{\tabcolsep}{5pt}
\renewcommand{\arraystretch}{0.95}
\resizebox{\linewidth}{!}{
\begin{tabular}{c|ccc|ccc}
\toprule[1.5pt]
\multirow{2}{*}{\textbf{\makecell{Max\\Rounds}}}
& \multicolumn{3}{c|}{\textbf{Response Performance}}
& \multicolumn{3}{c}{\textbf{Process Statistics}} \\
\cline{2-7}
& \textbf{\makecell{Pass@1\\(\%)}}
& \textbf{\makecell{L2\\(\%)}}
& \textbf{\makecell{L3\\(\%)}}
& \textbf{\makecell{Avg. Tool\\Calls}}
& \textbf{\makecell{Avg.\\Rounds}}
& \textbf{\makecell{Tool Util.\\(\%)}}
\\
\midrule
$N=6$  & 16.17 & 27.27 & 12.33 & 5.27 & 5.84 & 87.8 \\
$N=8$  & 20.83 & 31.82 & 17.04 & 7.07 & 7.68 & \textbf{88.4} \\
$N=10$ & \textbf{26.33} & \textbf{40.26} & 21.52 & 8.83 & 9.45 & 88.3 \\
$N=12$ & 26.17 & 38.31 & \textbf{21.97} & 9.67 & 10.43 & 80.6 \\
$N=16$ & 25.50 & 37.66 & 21.30 & \textbf{12.03} & \textbf{15.31} & 75.2 \\
\bottomrule[1.5pt]
\end{tabular}
}
% \vspace{-15pt}
\end{table}

\subsection{In-Depth Analysis}
\label{sec:in_depth_analysis}

We study factors affecting MLLM performance on Visual DeepSearch tasks. Failure analysis is conducted on a sample of 320 failed \emph{Search+Crop} trajectories from a single experimental run. Three experts independently annotate each trajectory, assigning its earliest dominant failure according to a shared taxonomy. Inter-annotator agreement reaches Fleiss' $\kappa=0.81$. Table~\ref{tab:error_decomposition} summarizes the resulting distribution.

\textbf{(1) Wrong or missing visual anchors (27.5\%).}
Agents ground the query to an incorrect logo, object, person, or region, or fail to identify a usable anchor. Because retrieval depends on this decision, the error propagates through the trajectory.
\textbf{(2) Retrieval drift or unsupported evidence (24.1\%).}
Agents issue underspecified queries, follow similarly named entities, or retain evidence that does not support the intended relation. Repeated search then reinforces the wrong branch instead of correcting it.
\textbf{(3) Long-horizon planning collapse (21.6\%).}
Agents omit intermediate subgoals, lose track of verified entities, or terminate before completing the evidence chain. Without explicit state tracking or backtracking, retrieval failures cascade across steps, especially on L3 tasks.
\textbf{(4) Multi-anchor fusion or calculation errors (16.9\%).}
Agents may solve individual chains but associate values with the wrong anchors or apply an incorrect fusion operation.
\textbf{(5) Response normalization or judge boundary cases (10.0\%).}
Semantically compatible responses may use ambiguous aliases, units, dates, or levels of specificity.

\section{Conclusion}
We introduce \bench, a benchmark designed specifically to evaluate Visual DeepSearch with 600 Visual DeepSearch tasks spanning 5 categories and 25 visual search scenarios. \bench tests repeated image inspection, visual-anchor grounding, and long-horizon evidence traversal with external knowledge. We also develop \eval, a unified evaluation framework for direct response generation, search-augmented reasoning, and crop-assisted visual inspection. Experiments on seven MLLMs show that current models remain far from solving the benchmark: the best model, SenseNova-MARS-32B, achieves only 26.33\% Pass@1.

% The source manuscript currently omits a separate limitations section.
% \section*{Limitations}
% \input{sections/10-limitaion.tex}

% \bibliographystyle{ACM-Reference-Format}
\bibliography{citation.bib}

\begin{appendices}

\section{Visual DeepSearch Task Examples}
\label{app:task_examples}
\raggedbottom

This appendix presents representative L2 tasks from different visual domains,
illustrating how image-grounded clues lead to long-horizon evidence chains and
cross-domain target identification.

\begin{figure}[H]
    \centering
    \includegraphics[width=\linewidth]{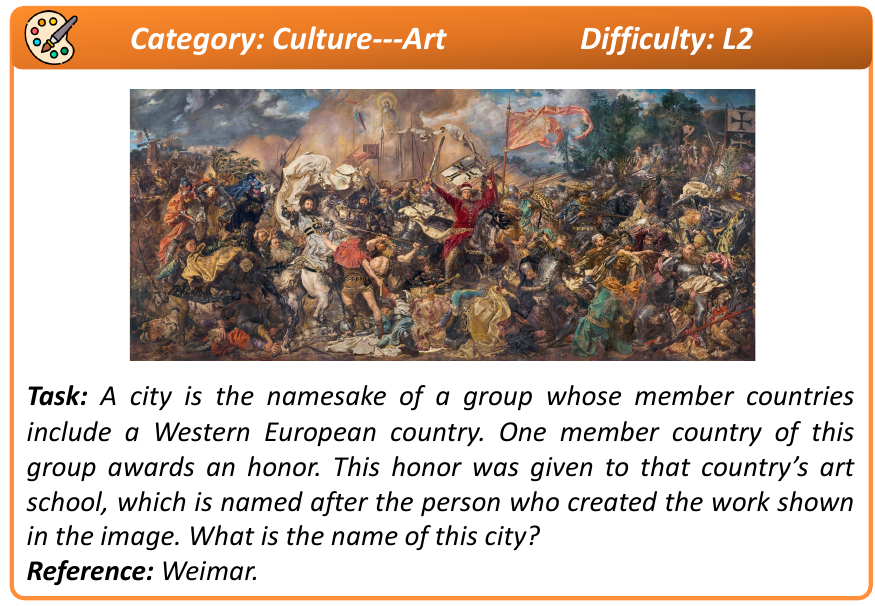}
    \caption{A long-horizon DeepSearch L2 art-domain example requiring painting identification and multi-hop reasoning to identify a city.}
    \label{fig:example_1}
    \vspace{-5pt}
\end{figure}

\begin{figure}[H]
    \centering
    \includegraphics[width=\linewidth]{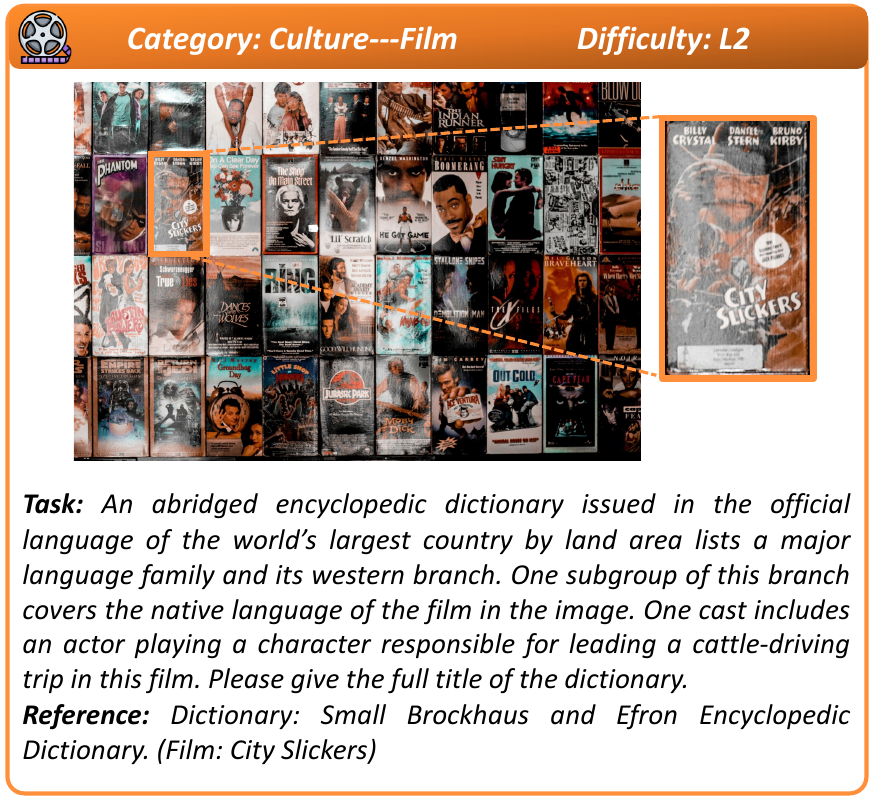}
    \caption{A long-horizon DeepSearch L2 film-domain example requiring localized poster recognition and cross-domain reasoning to identify a dictionary.}
    \label{fig:example_2}
    \vspace{-5pt}
\end{figure}

\begin{figure}[H]
    \centering
    \includegraphics[width=\linewidth]{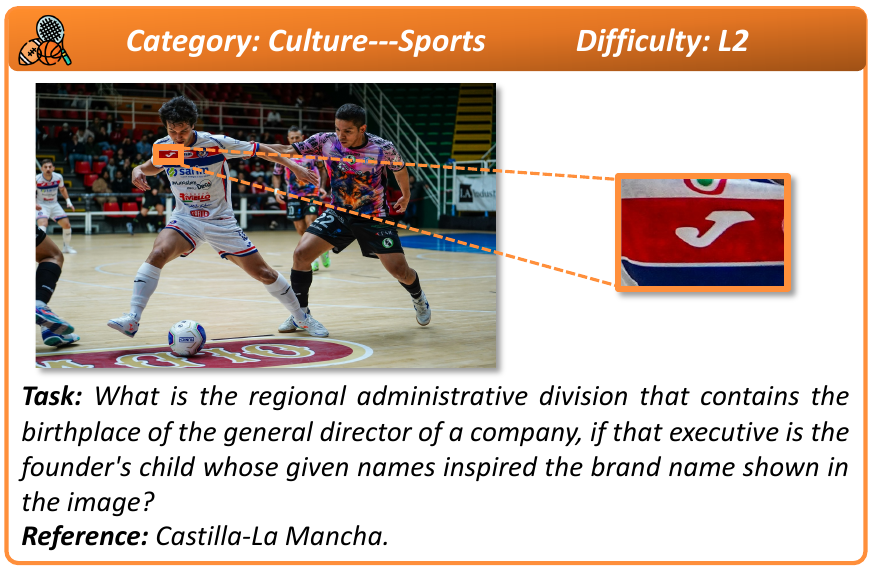}
    \caption{A long-horizon DeepSearch L2 sports-domain example requiring logo recognition and multi-hop reasoning to identify an administrative region.}
    \label{fig:example_3}
    \vspace{-5pt}
\end{figure}

\begin{figure}[H]
    \centering
    \includegraphics[width=\linewidth]{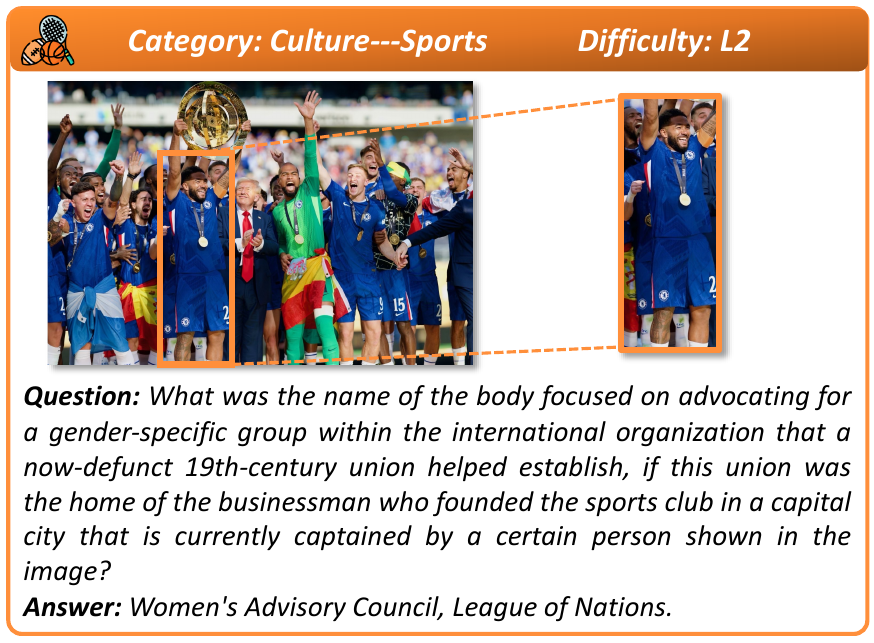}
    \caption{A long-horizon DeepSearch L2 sports-domain example requiring person identification and multi-hop reasoning to identify an international organization.}
    \label{fig:example_4}
    \vspace{-5pt}
\end{figure}

\begin{figure}[H]
    \centering
    \includegraphics[width=\linewidth]{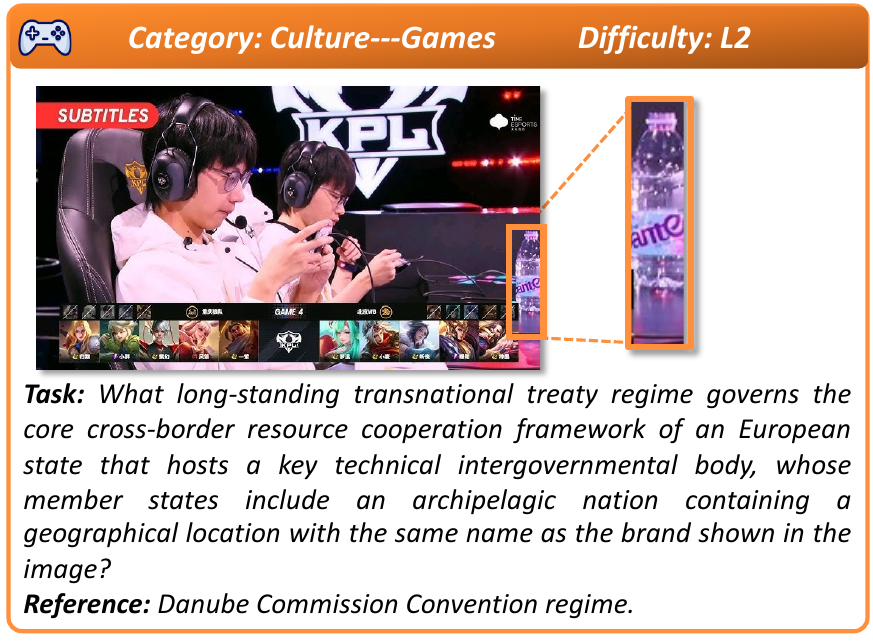}
    \caption{A long-horizon DeepSearch L2 games-domain example requiring fine-grained brand recognition and multi-hop reasoning to identify a treaty regime.}
    \label{fig:example_5}
    \vspace{-5pt}
\end{figure}

\begin{figure}[H]
    \centering
    \includegraphics[width=\linewidth]{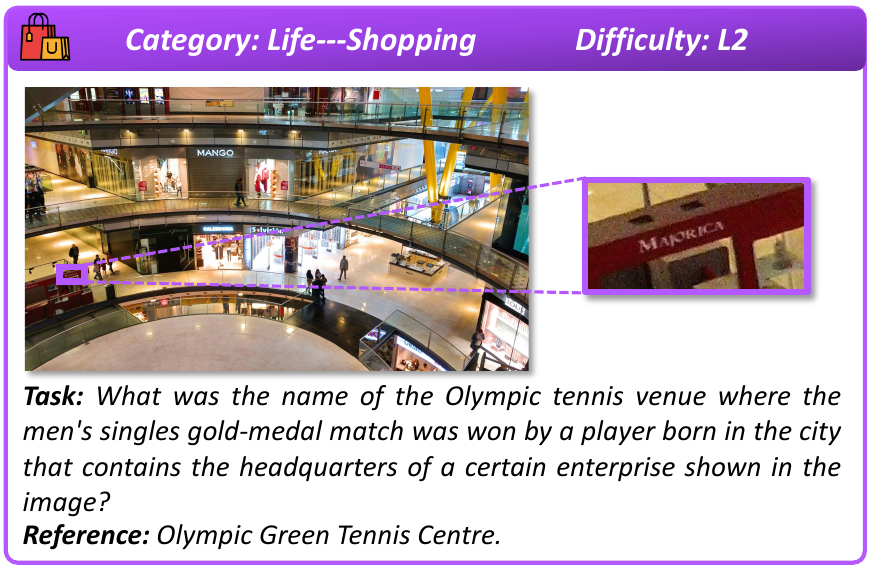}
    \caption{A long-horizon DeepSearch L2 shopping-domain example requiring storefront recognition and multi-hop reasoning to identify an Olympic venue.}
    \label{fig:example_6}
    \vspace{-5pt}
\end{figure}

\begin{figure}[!t]
    \centering
    \includegraphics[width=\linewidth]{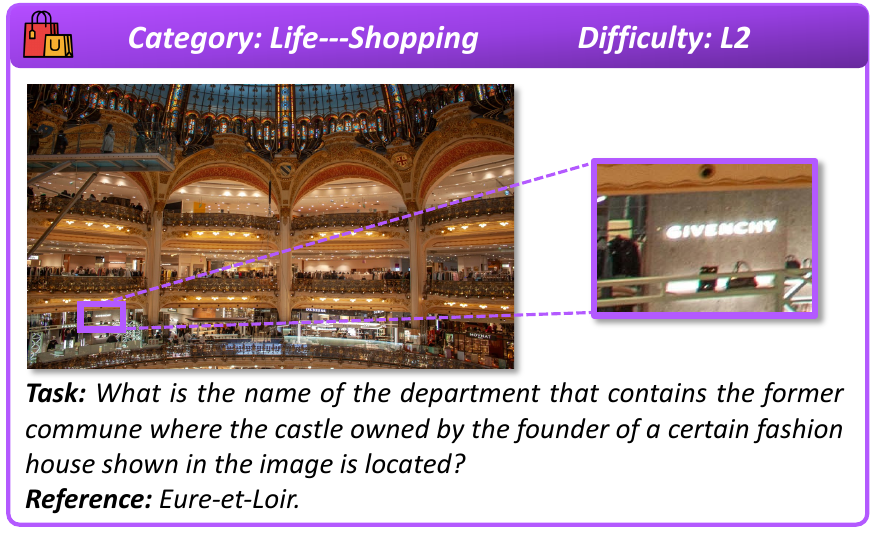}
    \caption{A long-horizon DeepSearch L2 shopping-domain example requiring fashion-brand recognition and multi-hop reasoning to identify an administrative department.}
    \label{fig:example_7}
    \vspace{-5pt}
\end{figure}

\definecolor{promptaccent}{HTML}{496A83}
\definecolor{promptborder}{HTML}{C7D1D9}
\definecolor{prompttitlebg}{HTML}{E9EFF3}
\definecolor{promptbg}{HTML}{F8FAFB}
\definecolor{promptink}{HTML}{263746}

\newenvironment{prompttablebox}[3]{%
\phantomsection\label{#2}%
\begin{tcolorbox}[
  enhanced,
  breakable,
  width=\columnwidth,
  colback=promptbg,
  colframe=promptborder,
  colbacktitle=prompttitlebg,
  coltitle=promptink,
  boxrule=0.35pt,
  borderline west={1.25pt}{0pt}{promptaccent},
  arc=1.2pt,
  outer arc=1.2pt,
  left=4.5pt,
  right=4.5pt,
  top=4pt,
  bottom=4pt,
  toptitle=2.8pt,
  bottomtitle=2.8pt,
  before skip=5pt,
  after skip=8pt,
  title={#3},
  fonttitle=\sffamily\bfseries\fontsize{7.4}{8.6}\selectfont,
  fontupper=\ttfamily\fontsize{6.55}{8.15}\selectfont\color{black!88},
  before upper={\setlength{\parindent}{0pt}\setlength{\parskip}{0.28em}\setlength{\emergencystretch}{1em}\raggedright}
]
}{%
\end{tcolorbox}
}

\section{Quality Control, Anti-Leakage Filtering, and Dataset Composition}
\label{app:qc_filtering_composition}
\subsection{Quality-Control and Anti-Leakage Results}

\paragraph{Sensitivity to construction-model choice.}
\label{app:construction_model_sensitivity}
During pipeline development, we tested several strong VLMs at different
stages, including entity extraction and OCR, query generation, and relation
verification. These models produced modestly different candidate outputs. The
difference was most apparent in entity extraction, where models varied in
their sensitivity to particular entity types and therefore proposed somewhat
different entity sets. We treat these model outputs only as candidate
proposals rather than as ground-truth annotations. Before an extracted entity
is allowed to seed a reasoning chain, human annotators verify its localized
visual evidence, identity or OCR result, and whether it can support a valid
extension of the evidence chain. All subsequent candidates, regardless of the
model that generated them, are subjected to the same relation, chain,
anti-leakage, and final human-validity checks. Consequently, construction-model
choice can affect the coverage and yield of the initial candidate pool, but
its influence on the correctness of the retained benchmark is substantially
attenuated by the shared verification and acceptance criteria.

\begin{table}[htbp]
\centering
\scriptsize
\caption{Primary failure types in 320 incorrect \emph{Search+Crop} traces. Each trace is assigned its earliest dominant failure.}
\label{tab:error_decomposition}
\setlength{\tabcolsep}{4pt}
\begin{tabularx}{\linewidth}{X c}
\toprule[1.2pt]
\textbf{Failure Type} & \textbf{Share} \\
\midrule
Wrong or missing visual anchor & 27.5\% \\
Retrieval drift or unsupported external evidence & 24.1\% \\
Long-horizon planning collapse & 21.6\% \\
Multi-anchor fusion or calculation error & 16.9\% \\
Response normalization or judge boundary case & 10.0\% \\
\bottomrule[1.2pt]
\end{tabularx}
\vspace{-8pt}
\end{table}

\begin{table*}[t]
\centering
\scriptsize
\caption{Human quality control of task queries before and after revision.}
\label{tab:chain_audit}
\setlength{\tabcolsep}{5pt}
\renewcommand{\arraystretch}{1.08}
\begin{tabularx}{\textwidth}{>{\raggedright\arraybackslash}X >{\centering\arraybackslash}p{0.12\textwidth} >{\raggedright\arraybackslash}p{0.39\textwidth}}
\toprule[1.2pt]
\rowcolor{black!10}\multicolumn{3}{l}{\textbf{Panel A. Pre-revision structural quality}} \\
\rowcolor{black!4}\textbf{Check} & \textbf{Pass rate} & \textbf{Main failure mode} \\
Visual seed is correctly grounded in the image & 94.8\% & Ambiguous logo or text region \\
Adjacent relation is factually valid & 92.6\% & Unsupported or overly broad relation \\
Target answer is unique and normalized & 96.1\% & Alias or date-format ambiguity \\
Chain requires all annotated evidence steps & 89.7\% & Skippable intermediate entity \\
Query preserves the intended evidence chain & 91.3\% & Over-compressed clue wording \\
\bottomrule[1.2pt]
\end{tabularx}

\vspace{5pt}

\begin{tabularx}{\textwidth}{>{\raggedright\arraybackslash}X >{\centering\arraybackslash}p{0.12\textwidth} >{\centering\arraybackslash}p{0.10\textwidth} >{\raggedright\arraybackslash}p{0.34\textwidth}}
\toprule[1.2pt]
\rowcolor{black!10}\multicolumn{4}{l}{\textbf{Panel B. Post-revision query naturalness}} \\
\rowcolor{black!4}
\textbf{Criterion} & \textbf{Mean $\pm$ SD} & \textbf{Pass rate} & \textbf{Main issue} \\
Grammatical fluency & $4.56 \pm 0.48$ & 97.1\% & Grammar or awkward collocation \\
Syntactic clarity & $4.41 \pm 0.57$ & 94.0\% & Nested syntax or unclear reference \\
Pragmatic naturalness & $4.28 \pm 0.64$ & 90.6\% & Artificial information need \\
Cross-clue coherence & $4.31 \pm 0.61$ & 91.4\% & Mechanical clue listing \\
Conciseness & $4.25 \pm 0.66$ & 89.7\% & Redundancy or excess complexity \\
\midrule
Overall naturalness & $4.27 \pm 0.47$ & 91.1\% & Any failed component criterion \\
\quad Single-chain ($n=138$) & $4.48 \pm 0.39$ & 95.1\% & --- \\
\quad Multi-chain fusion ($n=462$) & $4.21 \pm 0.48$ & 89.9\% & --- \\
\bottomrule[1.2pt]
\end{tabularx}

\vspace{3pt}
\parbox{0.98\textwidth}{\scriptsize\emph{Note:} Panel A measures structural validity, not residual errors or linguistic naturalness in the retained benchmark. Panel B uses a 1--5 Likert scale. A criterion passes at a mean annotator score of at least 3; passing the overall check additionally requires a five-criterion mean of at least 4.}
\end{table*}

\paragraph{Human evaluation of query naturalness.}
Three annotators independently evaluate all 600 retained queries after human
revision. For each item, annotators see only the image and the final query; they
do not see the target, annotated evidence chain, generation history, or other
annotators' ratings. Items are presented in a randomized order, and the
annotators are not told whether an item is single-chain or multi-chain. Each
criterion is rated on a five-point Likert scale, where 1 denotes a severe
problem, 3 denotes acceptable language with noticeable defects, and 5 denotes
fully natural language.

The rubric separates five aspects of query quality. \emph{Grammatical fluency}
covers grammar, lexical choice, collocation, and sentence-level transitions.
\emph{Syntactic clarity} measures ease of parsing, penalizing deeply nested
modifiers and unclear antecedents. \emph{Pragmatic naturalness} asks whether the
query resembles a plausible information-seeking request rather than a
benchmark-specific construction. \emph{Cross-clue coherence} assesses whether
the evidence-bearing clues form a connected question instead of a mechanical
list. \emph{Conciseness} penalizes repetition, avoidable qualifications, and
unnecessary syntactic complexity. We define overall naturalness as the mean of
these five scores. A query passes the overall naturalness check only when every
criterion score is at least 3 and its overall mean is at least 4.

For each criterion, we report the mean, standard deviation, criterion-level
pass rate, and inter-annotator agreement measured by Krippendorff's $\alpha=0.82$ for
ordinal ratings. We additionally report overall results separately for single-chain and multi-chain queries, since fusion may introduce longer
sentences, denser clue stacking, and more list-like phrasing. Panel A of
Table~\ref{tab:chain_audit} diagnoses errors in the pre-revision candidate pool,
including faulty visual grounding, unsupported relations, non-unique targets,
skippable evidence steps, and loss of the intended chain. Panel B instead
measures the language quality of the retained queries. A query may therefore
preserve its intended evidence chain while still receiving a low naturalness
score because of nested syntax, unclear references, or mechanical clue
composition.

\begin{table}[htbp]
\centering
\scriptsize
\caption{Filtering statistics for candidate items entering each corresponding check. Rewritten items are retained only after passing the next verification round; the final text-only row is an aggregate diagnostic rather than a per-item leakage rate.}
\label{tab:leakage_stats}
\setlength{\tabcolsep}{4pt}
\begin{tabularx}{\linewidth}{X c c}
\toprule[1.2pt]
\textbf{Stage} & \textbf{Affected} & \textbf{Action} \\
\midrule
Text-only solver finds shortcut & 27.1\% & Rewrite \\
Unresolved after $R_{\max}=3$ rounds & 8.5\% & Discard \\
Clue-necessity verification finds redundancy & 18.6\% & Shorten/recompute/discard \\
Low visual-dependence score & 7.7\% & Inspect/discard \\
Human annotators mark ambiguous & 6.9\% & Revise/discard \\
Final no-image text-only accuracy & 2.7\% & Aggregate diagnostic \\
\bottomrule[1.2pt]
\end{tabularx}
\vspace{-6pt}
\end{table}

\paragraph{Visual-dependence threshold selection and sensitivity.}
\label{app:visual_dependence_threshold}
The visual-dependence score is a difference in target-sequence log likelihoods.
Consequently, exponentiating a cutoff $\tau$ gives its likelihood-ratio
interpretation:
\begin{equation}
\exp(\tau)
=
\frac{\pi_{\theta}(t_j\mid\tilde{q}_j,I_j)}
{\pi_{\theta}(t_j\mid\tilde{q}_j,I_{j,\mathrm{mask}})}.
\end{equation}
We use $\tau=0.182$, for which $\exp(\tau)=1.1996$, requiring the target sequence to be
1.1996 times as likely under the original image as under its masked
counterpart. A zero cutoff would require only a positive difference and would
therefore admit arbitrarily small changes in likelihood, while a cutoff of 0.1
corresponds to a likelihood ratio of $\exp(0.1)=1.1052$. Table~\ref{tab:visual_threshold_interpretation}
shows the implied margins for several candidate thresholds.

\begin{table}[htbp]
\centering
\scriptsize
\caption{Interpretation of candidate visual-dependence cutoffs. The likelihood ratio is $\exp(\tau)$.}
\label{tab:visual_threshold_interpretation}
\setlength{\tabcolsep}{4pt}
\begin{tabularx}{\linewidth}{c c X}
\toprule[1.2pt]
\textbf{Cutoff $\tau$} & \textbf{Likelihood ratio} & \textbf{Interpretation} \\
\midrule
0.000 & 1.00 & No positive likelihood margin \\
0.100 & 1.11 & Weak positive margin \\
0.182 & 1.20 & Selected threshold used in \bench \\
0.300 & 1.35 & More conservative screen \\
0.500 & 1.65 & Substantially more conservative screen \\
\bottomrule[1.2pt]
\end{tabularx}
\vspace{-6pt}
\end{table}

The cutoff is a triage rule rather than an automatic acceptance criterion.
Items at or below 0.182 are routed to adjudicated review, and items above it
must still pass the remaining grounding, clue-necessity, anti-leakage, and human
verification checks. For this reason, the Qwen3-VL-32B-Instruct score is not
treated as a binary ground-truth classifier whose output alone determines
benchmark membership. Taking visually necessary items as the positive class,
the scorer achieved 95.0\% precision and 92.3\% recall at this threshold,
corresponding to an F1 score of 93.6\%. These measured figures
characterize the observed screening behavior of the scorer rather than
an automatic acceptance rule. The value 0.182 is also scorer specific: likelihood
calibration may differ across architectures, tokenizers, and target lengths.
Applying the procedure with another scorer therefore requires either preserving
the same likelihood-ratio interpretation or selecting a scorer-specific operating
point, together with adjudication of borderline cases. Claims of visual necessity
rest on the complete verification procedure rather than on transfer of the Qwen
score to other models.

\paragraph{Scorer robustness and proprietary-model access.}
As a robustness check, we also repeated this diagnostic with other open-weight
multimodal scorers, including the larger Qwen3-VL-235B-A22B-Instruct
model~\cite{qwen2025qwen3vl235ba22binstruct}. The larger scorer produced broadly
similar triage outcomes and did not materially improve separation from human
adjudication, indicating that model scale alone does not make this diagnostic
more reliable. We could not include proprietary models available only through
inference APIs because the score requires the conditional log likelihood of an
arbitrary fixed target sequence under both the original and masked images, rather
than the probability of a sampled response. The APIs available to us did not
expose reproducible token-level likelihoods for teacher-forced targets with visual
inputs, or sufficient control over model version and image preprocessing to
compute the paired score as defined above.

The final no-image text-only accuracy is measured at the model level over the retained set. It is not interpreted as the percentage of leaking items because isolated correct responses can arise from guessing or parametric recall. We therefore claim that the retained benchmark is not \emph{reliably} solvable from query text alone, rather than that every possible text-only solver must fail on every item.

\subsection{Dataset Composition}

Table~\ref{tab:dataset_composition} provides a detailed breakdown of the 600
retained tasks by answer type, visual-anchor type, and reasoning form. The
benchmark covers diverse target formats, including entity names, dates,
numeric values, locations, and comparative or selection-based answers. Its
visual anchors likewise span organization logos, visible text, products,
landmarks, people, events, and symbols, reducing dependence on any single
recognition capability. In terms of reasoning structure, 138 tasks contain a
single evidence chain, while the remaining 462 require multi-chain fusion.
Among the fused tasks, 203 combine two component chains, 252 combine three,
six combine four, and one combines six. These distributions
show that \bench evaluates not only long-horizon evidence traversal, but also
heterogeneous visual grounding and target derivation.

\begin{table}[htbp]
\centering
\scriptsize
\caption{Dataset composition beyond category-level counts. The table separates sources of difficulty that are otherwise conflated in aggregate accuracy.}
\label{tab:dataset_composition}
\setlength{\tabcolsep}{2.5pt}
\begin{tabularx}{\linewidth}{p{0.31\linewidth} X c}
\toprule[1.2pt]
\textbf{Dimension} & \textbf{Type} & \textbf{\#} \\
\midrule
\multirow{5}{*}{Answer type} 
& Entity/name & 243 \\
& Date/year & 96 \\
& Count or numeric value & 110 \\
& Location & 70 \\
& Comparative/selection & 81 \\
\midrule
\multirow{5}{*}{Visual anchor} 
& Organization/logo & 177 \\
& Text/signage/OCR & 122 \\
& Product/object & 106 \\
& Landmark/place & 92 \\
& Person/event/symbol & 103 \\
\midrule
\multirow{5}{*}{Reasoning form}
& Single evidence chain & 138 \\
& Two-chain fusion & 203 \\
& Three-chain fusion & 252 \\
& Four-chain fusion & 6 \\
& Six-chain fusion & 1 \\
\bottomrule[1.2pt]
\end{tabularx}
\vspace{-6pt}
\end{table}

As shown in Table~\ref{tab:dataset_composition}, the counts within each
dimension sum to 600. The four fusion categories sum to 462 and constitute a
mutually exclusive breakdown of the multi-chain subset, whereas the 138
single-evidence-chain tasks form the single-chain subset.

\paragraph{Entity-type coverage.}
The visual-anchor panel of Table~\ref{tab:dataset_composition} also reports the
task-level distribution of the primary image-grounded entity family. We use
the primary anchor rather than all candidate regions produced during image
segmentation because only the former is retained as task-relevant evidence.
Organization and logo anchors account for 177 tasks (29.5\%), followed by
text, signage, or OCR-bearing entities (122; 20.3\%), products or objects
(106; 17.7\%), people, events, or symbols (103; 17.2\%), and landmarks or
places (92; 15.3\%). These categories are mutually exclusive at the task level.
A task may nevertheless contain additional entities, and multi-chain tasks may
ground different component chains in different image regions.

\begin{figure}[htbp]
\centering
\includegraphics[width=0.8\linewidth]{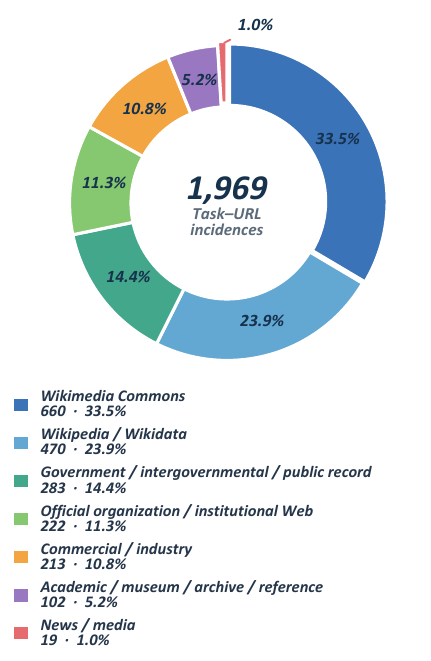}
\caption{Source-domain distribution over 1,969 Task--URL evidence incidences.
URLs are deduplicated within each task but may be counted for multiple tasks;
the legend reports both the incidence count and corpus share.}
\label{fig:source_domain_distribution}
\vspace{-6pt}
\end{figure}

\paragraph{Source-domain coverage.}
Figure~\ref{fig:source_domain_distribution} reports the provenance of the
evidence used by the retained tasks. We canonicalize each URL to its host,
remove duplicate URLs within an item, and then count Task--URL incidences; the
same source used by two different tasks therefore contributes two incidences.
The 600 tasks contain 1,969 such incidences across 554 distinct hosts.
Wikimedia Commons and Wikipedia/Wikidata remain the two largest families,
reflecting the benchmark's emphasis on auditable image provenance and linkable
entities. Importantly, 42.6\% of the evidence incidences come from outside
these two families, including government and intergovernmental records,
official institutional pages, academic and cultural collections, news
organizations, and commercial or industry sources. Source families are
assigned from the canonical host and publisher identity; government,
intergovernmental, academic, and cultural institutions take precedence over a
generic top-level-domain rule.

\paragraph{Relation-type coverage.}
Table~\ref{tab:relation_type_distribution} gives the distribution of the seven
normalized semantic relation classes used during evidence-chain construction.
We count each directed edge occurrence rather than each unique predicate,
because repeated uses of a relation contribute separately to the reasoning
load. We audit all 600 released construction records and select one
authoritative edge sequence per component chain, avoiding duplicate
representations of the same edge in nested fields. Because the five
construction batches use different schema versions, explicit normalized
labels are canonicalized first; records that retain only a raw predicate or
an evidence-edge role are assigned by a deterministic lexical crosswalk to
the same seven-class inventory. Quantitative property retrieval is treated as
attributive, event and history links as temporal, and location and route links
as spatial. This yields 9,008 normalized edge occurrences across the complete
600-task corpus.

\begin{table}[htbp]
\centering
\scriptsize
\caption{Normalized semantic-relation distribution for all 9,008 recorded
edge occurrences in the complete 600-task corpus; percentages are rounded
independently.}
\label{tab:relation_type_distribution}
\setlength{\tabcolsep}{4pt}
\begin{tabularx}{\linewidth}{X r r}
\toprule[1.2pt]
\textbf{Relation type} & \textbf{\# edges} & \textbf{\%} \\
\midrule
Attributive & 4,857 & 53.92 \\
Member--collection & 1,587 & 17.62 \\
Comparative & 1,071 & 11.89 \\
Temporal & 523 & 5.81 \\
Part--whole & 462 & 5.13 \\
Spatial & 452 & 5.02 \\
Causal & 56 & 0.62 \\
\midrule
Total & 9,008 & 100.00 \\
\bottomrule[1.2pt]
\end{tabularx}
\vspace{-6pt}
\end{table}

\subsection{Benchmark Feature Audit Protocol}
\label{appendix:benchmark_feature_audit}

Table~\ref{tab:benchmark_comparison} is intended as a structured feature audit,
not as a qualitative ranking of prior benchmarks. We assess each entry using
the corresponding benchmark paper, released task schema, and evaluation
protocol. A feature is marked as fully addressed ({\cmark}) only when it is an
explicit benchmark-wide requirement supported by the released annotations or
evaluation procedure. It is marked as partially addressed ({\notcheckmark}) when
it is encouraged, indirectly supported, or present in only a subset of tasks,
but is not systematically enforced and evaluated. It is marked as not addressed
({\xmark}) when the reviewed public materials provide no documented mechanism for
enforcing or evaluating the feature. Thus, {\xmark} denotes an absence of
documented benchmark-level support, rather than proof that the feature never
occurs in any individual instance.

We apply the following operational criteria. \emph{Vision-centric search}
requires the answer to depend on a visually grounded search clue rather than
using the image only as optional context. \emph{Fine-grained visual retrieval}
requires localization or retrieval involving an image region, object, visible
text, or visual attribute. \emph{Repeated image inspection} requires visual
evidence to be acquired at multiple stages of the intended solution rather than
only once at the beginning. \emph{Deep long-horizon search} requires annotated
multi-step evidence traversal, including tasks that satisfy the L3 threshold
defined in Table~\ref{tab:difficulty_levels}. \emph{Temporal validity} requires
timestamps, an explicit validity period, periodic refresh, or a revalidation
mechanism for time-sensitive targets. \emph{Target uniqueness} requires an
evidence-backed uniqueness check, answer normalization, or human adjudication
of ambiguity.

The annotated anchors and chains, together with the evaluator's tool-call trace,
provide a basis for extending the evaluation, but the main results currently
score final answers and report only aggregate interaction statistics. Region
masks or boxes and explicit
anchor--evidence binding labels must additionally be released for process-scored
runs. Multiple annotated anchors do not guarantee temporally interleaved visual
access: a model may recognize all anchors during its initial full-image
observation and perform the remaining steps through text search alone. The
protocol below states what must be measured before making a model-level
repeated-inspection claim.

\subsection{Process-Level Grounding Metrics and Repeated-Inspection Controls}
\label{appendix:repeated_inspection_protocol}

\paragraph{Scope.}
The ablation in Table~\ref{tab:model_performance} separates the marginal utility
of external search and cropping, but it does not identify which image regions
were used or whether visual evidence was revisited. In particular, the
4.80-point average gain from \emph{No-Tool} to \emph{Search} is smaller than the
additional 8.39-point gain from \emph{Search} to \emph{Search+Crop}. This result
shows that external retrieval is beneficial, while localized inspection
provides the larger marginal improvement and remains a major bottleneck. To
characterize how agents acquire, revisit, and connect visual evidence, we extend
\eval with the process-level metrics and controlled evaluations below.

\paragraph{Reference and trace representation.}
For item $i$, let
$\mathcal{A}_i=\{(r_{ij},e_{ij},c_{ij})\}_{j=1}^{m_i}$ denote its necessary
visual anchors, where $r_{ij}$ is the reference mask or bounding box,
$e_{ij}$ is the grounded entity, and $c_{ij}$ is the component-chain
identifier. Let $\mathcal{G}_i=(\mathcal{V}_i,\mathcal{E}_i)$ be the verified
evidence graph. A process-scored run stores a time-ordered trace containing
every image region viewed, the entity asserted for that region, every retrieved
source and atomic claim, and the anchor or component chain to which the model
assigns that claim. This structured record is used only for evaluation and is
not populated from the reference chain during inference.

A predicted anchor matches a reference anchor only when (i) the predicted
entity is an accepted alias or is judged entity-equivalent and (ii) its region
has intersection-over-union of at least 0.5 with the reference region. For
coverage metrics, a crop covers an anchor when it contains at least 50\% of the
anchor area. To prevent a second full-image view from receiving localization
credit, an eligible local interaction must cover no more than 50\% of the input
image. We report sensitivity to these two thresholds. Duplicate crops and
paraphrases are merged before scoring.

\paragraph{Visual grounding.}
Let $M_i$ be the maximum one-to-one matching between predicted and reference
anchors. We report macro-averaged per-item precision and recall:
\begin{equation}
\operatorname{VA\text{-}P}_i
=\frac{|M_i|}{|\widehat{\mathcal{A}}_i|},
\qquad
\operatorname{VA\text{-}R}_i
=\frac{|M_i|}{|\mathcal{A}_i|}.
\end{equation}
An empty prediction has zero precision and recall. \emph{Necessary Visual Region
Coverage} (NVRC) is the fraction of reference anchors covered by at least one
eligible local interaction, irrespective of the asserted identity:
\begin{equation}
\operatorname{NVRC}_i
=\frac{1}{m_i}\sum_{j=1}^{m_i}
\mathbf{1}\!\left[
\exists\,\widehat r\in\tau_i:
\frac{|r_{ij}\cap\widehat r|}{|r_{ij}|}\geq 0.5
\right].
\end{equation}
VA-R measures whether the correct visual entities were recovered, whereas NVRC
separates localization from naming accuracy.

\paragraph{Evidence-chain recovery.}
We atomize the trajectory into entity or attribute nodes and directed relation
claims, then align them to the minimum sufficient reference graph. Node
Coverage and Relation Coverage are
\begin{equation}
\operatorname{NC}_i
=\frac{|\widehat{\mathcal{V}}_i\cap\mathcal{V}_i^{\rm req}|}
{|\mathcal{V}_i^{\rm req}|},
\qquad
\operatorname{RC}_i
=\frac{|\widehat{\mathcal{E}}_i\cap\mathcal{E}_i^{\rm req}|}
{|\mathcal{E}_i^{\rm req}|}.
\end{equation}
Matching requires entity equivalence and the correct relation direction; merely
mentioning both endpoint entities is insufficient. Because a valid agent may
find a different evidential route, an alternative path receives credit only
after source-backed adjudication confirms that it is sufficient, non-circular,
and preserves target uniqueness. Scores are reported over all trajectories,
not only those with a correct final answer.

\paragraph{Evidence grounding and anchor binding.}
For every external atomic claim $z$, annotators or a calibrated entailment
checker determine whether the cited source supports $z$. \emph{Evidence
Accuracy} (EA) is the fraction of submitted external claims that are supported.
\emph{Anchor--Evidence Binding Accuracy} (AEBA) is the fraction of supported
claims assigned to the correct visual anchor and component chain.
\emph{Grounded Evidence Accuracy} (GEA) applies both requirements jointly. Let
$n_i$ be the number of submitted external claims, $s_i$ the number supported by
their cited sources, and $b_i$ the number that are both supported and bound to
the correct anchor and chain. Then
\begin{equation}
\operatorname{EA}_i=\frac{s_i}{n_i},\qquad
\operatorname{AEBA}_i=\frac{b_i}{s_i},\qquad
\operatorname{GEA}_i=\frac{b_i}{n_i}.
\end{equation}
Metrics with an empty denominator are defined as zero.
Claims without a source or an explicit anchor/chain assignment receive no
grounding credit. This prevents a correct retrieved fact attached to the wrong
logo, person, or component from being counted as a correct process step.

\paragraph{Repeated and distinct visual interaction.}
We report \emph{Distinct Region Interaction} (DRI), the number of distinct
necessary anchor regions receiving an eligible local interaction, together with
its normalized form $\operatorname{DRI}_i/m_i$. A \emph{revisit} is stricter:
the agent must inspect a reference anchor, perform at least one intervening
evidence-producing external search, and subsequently inspect that anchor or
another still-unresolved necessary anchor. Immediate duplicate crops do not
count. Revisit Rate is the fraction of tasks with at least one such interleaved
visual return; we additionally report the mean number of valid revisits and the
round of the first revisit. We separately report same-anchor revisits and
returns to a different unresolved anchor. These definitions distinguish
repeated visual evidence seeking from front-loaded multi-anchor recognition.

\begin{table}[htbp]
\centering
\scriptsize
\caption{Matched controls for isolating repeated visual inspection. Text
retrieval snapshots, decoding settings, and answer scoring are held fixed.}
\label{tab:repeated_inspection_controls}
\setlength{\tabcolsep}{3.5pt}
\begin{tabularx}{\linewidth}{p{0.25\linewidth} X}
\toprule[1.2pt]
\textbf{Condition} & \textbf{Visual-access intervention} \\
\midrule
Full-image only & Present the full image once in the initial turn; disable all subsequent image access and cropping. \\
Front-loaded anchors & During the initial observation, require a structured list of all proposed anchor regions and identities; then disable vision. This tests whether multiple anchors can be acquired at once. \\
Dynamic Crop & Present the same initial image and permit crops from the original image between retrieval steps. \\
Necessary-region mask & Mask each necessary anchor separately and all necessary anchors jointly while preserving the rest of the image. \\
Matched distractor & Replace a necessary region with a size- and salience-matched distractor while leaving the query and cached text-retrieval snapshot unchanged. \\
\bottomrule[1.2pt]
\end{tabularx}
\vspace{-6pt}
\end{table}

\paragraph{Budget matching and causal contrasts.}
The full-image and dynamic-crop conditions use the same model, prompts,
decoding seeds, maximum rounds, cached retrieval results, and text-search
budget. Visual-action slots are reserved in advance: when a crop is unavailable
in a control condition, its slot yields a null observation and cannot be
converted into an additional text search. Thus, any difference is not explained
by unequal access to external text evidence. We report paired differences in
Pass@1, VA-R, NVRC, RC, GEA, and AEBA with bootstrap confidence intervals,
separately for single-chain and multi-chain items.

Table~\ref{tab:process_grounding_results} reports this controlled diagnostic
using SenseNova-MARS-32B. Absolute metric values are reported for all three
visual-access conditions. The $\Delta$ columns contain the paired difference
between \emph{Dynamic Crop} and \emph{Full-image only}, together with a 95\%
bootstrap confidence interval over items.

\begin{table}[htbp]
\centering
\scriptsize
\caption{Process-level grounding and repeated-inspection results for
SenseNova-MARS-32B. F denotes \emph{Full-image only}, A denotes
\emph{Front-loaded anchors}, and D denotes \emph{Dynamic Crop}.
$\Delta=\mathrm{D}-\mathrm{F}$; values in brackets are paired 95\% bootstrap
confidence intervals. Higher is better for all metrics.}
\label{tab:process_grounding_results}
\setlength{\tabcolsep}{2.5pt}
\renewcommand{\arraystretch}{1.05}
\begin{tabularx}{\linewidth}{>{\raggedright\arraybackslash}X c c c c}
\toprule[1.2pt]
\textbf{Metric} & \textbf{F} & \textbf{A} & \textbf{D}
& \textbf{$\Delta$ [95\% CI]} \\
\midrule
\multicolumn{5}{l}{\textbf{Panel A. Single-chain}} \\
\midrule
Pass@1 (\%)                & 29.13 & 34.49 & 41.16 & 12.03 [5.80, 18.14] \\
Visual Anchor Recall (\%)  & 51.74 & 60.87 & 68.26 & 16.52 [9.46, 23.33] \\
NVRC (\%)                  & 0.00 & 0.00 & 62.32 & 62.32 [54.10, 70.05] \\
Relation Coverage (\%)     & 42.18 & 49.64 & 55.80 & 13.62 [8.33, 18.91] \\
GEA (\%)                   & 38.26 & 44.71 & 50.65 & 12.39 [6.96, 17.75] \\
AEBA (\%)                  & 53.91 & 63.48 & 69.57 & 15.66 [9.22, 21.90] \\
Normalized DRI (\%)        & 0.00 & 0.00 & 62.32 & 62.32 [54.10, 70.05] \\
Revisit Rate (\%)          & 0.00 & 0.00 & 31.88 & 31.88 [24.10, 40.11] \\
\midrule
\multicolumn{5}{l}{\textbf{Panel B. Multi-chain}} \\
\midrule
Pass@1 (\%)                & 15.00 & 18.42 & 21.90 & 6.90 [3.75, 10.08] \\
Visual Anchor Recall (\%)  & 35.84 & 44.37 & 50.54 & 14.70 [10.59, 18.76] \\
NVRC (\%)                  & 0.00 & 0.00 & 49.13 & 49.13 [44.59, 53.68] \\
Relation Coverage (\%)     & 29.42 & 34.81 & 39.76 & 10.34 [7.29, 13.41] \\
GEA (\%)                   & 25.76 & 31.15 & 35.93 & 10.17 [7.10, 13.25] \\
AEBA (\%)                  & 39.18 & 47.32 & 54.61 & 15.43 [11.64, 19.22] \\
Normalized DRI (\%)        & 0.00 & 0.00 & 49.13 & 49.13 [44.59, 53.68] \\
Revisit Rate (\%)          & 0.00 & 0.00 & 44.16 & 44.16 [39.65, 48.71] \\
\bottomrule[1.2pt]
\end{tabularx}
\vspace{-6pt}
\end{table}

For the counterfactual conditions, the primary quantities are
\begin{align}
\Delta_{\rm mask}
&=\operatorname{P@1}(I)-\operatorname{P@1}(I_{\rm mask}),\\
\Delta_{\rm dist}
&=\operatorname{P@1}(I)-\operatorname{P@1}(I_{\rm distractor}).
\end{align}

\begin{table}[htbp]
\centering
\scriptsize
\caption{Necessary-region counterfactual results for
SenseNova-MARS-32B under \emph{Dynamic Crop}. ``Original,'' ``Mask,'' and
``Distractor'' report Pass@1 (\%); both $\Delta$ columns are paired
performance drops with 95\% bootstrap confidence intervals.}
\label{tab:region_counterfactual_results}
\setlength{\tabcolsep}{2.5pt}
\renewcommand{\arraystretch}{1.05}
\resizebox{\linewidth}{!}{
\begin{tabular}{l c c c c c}
\toprule[1.2pt]
\textbf{Split}
& \textbf{Original}
& \textbf{Mask}
& \textbf{$\Delta_{\rm mask}$ [95\% CI]}
& \textbf{Distractor}
& \textbf{$\Delta_{\rm dist}$ [95\% CI]} \\
\midrule
Single-chain & 41.16 & 23.91 & 17.25 [10.80, 23.74] & 20.87 & 20.29 [13.53, 27.03] \\
Multi-chain  & 21.90 & 10.95 & 10.95 [7.89, 14.09] & 8.70 & 13.20 [9.93, 16.54] \\
Overall      & 26.33 & 13.93 & 12.40 [9.65, 15.20] & 11.50 & 14.83 [11.92, 17.78] \\
\bottomrule[1.2pt]
\end{tabular}
}
\vspace{-6pt}
\end{table}

Together, DRI and Revisit Rate measure repeated visual interaction,
region--entity grounding evaluates whether the correct image evidence is
inspected, and the necessary-region counterfactuals quantify the causal
contribution of that evidence. These process-level diagnostics complement
Pass@1 by separating final-answer correctness from visual grounding, evidence
revisiting, and anchor--evidence binding.

\subsection{Cross-Benchmark Hop Annotation Protocol}
\label{appendix:cross_benchmark_hops}

For the comparison in Figure~\ref{fig:baseline_difficulty_distribution}, we
define a hop as a task-relevant semantic transition that introduces a new
entity, attribute, relation, or intermediate conclusion required by a later
step or by the final answer. The initial transition from visual evidence to an
identified entity is therefore counted as a hop. By contrast, an operation is
not counted merely because it invokes a tool: preprocessing and navigation
actions such as cropping, zooming, query reformulation, and opening a webpage
do not constitute separate hops unless they themselves add new task-relevant
evidence. Repeated attempts that recover the same evidence are likewise counted
only once.

For benchmarks with native process annotations, we derive $H$ from the released
labels under this definition: annotated sub-goals for BrowseComp-$V^3$,
evidence-producing operations in the ground-truth tool chain for VTC-Bench, and
the released 3-hop/5-hop labels for the hop-labeled MTA-Agent subset. When a
released process annotation contains an auxiliary operation that does not
change the evidence state, that operation is not treated as an additional hop.
MMSearch and VDR-Bench do not provide per-instance hop annotations. For these
benchmarks, we ask an LLM to decompose each instance into a minimum sufficient
evidence chain connecting the visual input to the reference answer, following
the same evidence-transition definition. Human annotators then inspect every
generated decomposition, remove redundant or tool-only steps, add missing
evidence dependencies, and correct the ordering where necessary. The final
count $H$ is taken from the human-verified decomposition rather than from the
number of tool calls made by either the LLM or an evaluated agent. We map the
resulting counts to the common thresholds L1 ($H<5$), L2 ($5\leq H<10$), and
L3 ($H\geq10$).

% \clearpage
\section{Prompts in the Data Construction Pipeline}
\label{appendix:prompts}

This appendix lists the prompts used in the data construction pipeline described in Sections~\ref{sec:stage1}--\ref{sec:stage6}. 
The prompts are organized by pipeline stage. 
Variables enclosed in braces, such as \{entity\_name\}, \{processed\_width\}, and \{reasoning\_chain\_desc\}, are runtime placeholders filled by the corresponding generation script.

\subsection{Prompts for Image Source Filtering and Entity Extraction}
\label{appendix:prompts_stage1}

This stage extracts visually grounded named entities from each retained image and generates local visual descriptions for these entities.

\begin{prompttablebox}{Prompt for Stage 1 Pass 1 entity list extraction.}{tab:prompt_stage1_pass1}{Stage 1 --- Pass 1: Entity list extraction}
Analyze this image and identify distinctive, specific named entities that are visually grounded in the image. \par
\smallskip
\textbf{Image size.} \{processed\_width\}\(\times\)\{processed\_height\} pixels. \par
\smallskip
\textbf{Criteria.} \par
- It must be a specific named entity (person, brand, product, landmark, road, street, etc.). \par
- Must be something you could take a photo of \par
- Must be visually distinctive in this image \par
\smallskip
\textbf{Suitable examples.} \par
- Albert Einstein \par
- Toyota \par
- Eiffel Tower \par
- iPhone \par
- Statue of Liberty \par
- Hollywood Boulevard \par
- Fifth Avenue \par
- Route 66 \par
\smallskip
\textbf{Examples to avoid.} \par
- Person (use specific name if known) \par
- Car (use brand/model if visible) \par
- Building (use specific name if known) \par
- Tree \par
- Generic object \par
\smallskip
\textbf{Output format.} \par
Return a JSON array. Each entity must contain: \par
- raw\_label: the visible text or initial visual label \par
- canonical\_name: the normalized entity name \par
- entity\_type: PERSON/ORGANIZATION/PRODUCT/LOCATION/SYMBOL/EVENT \par
- bbox: [x1, y1, x2, y2] in pixel coordinates \par
- confidence: a value from 0.0 to 1.0 \par
- visual\_attributes: local appearance, color, text, logo pattern, or scene context supporting the identification \par
Example: \par
[\{"raw\_label": "CAMRY", "canonical\_name": "Toyota Camry", "entity\_type": "PRODUCT", "bbox": [120, 210, 360, 420], "confidence": 0.94, "visual\_attributes": "white sedan with a CAMRY badge on the rear"\}] \par
\smallskip
Return only entities that are supported by a clearly localized visual region. If the canonical identity is uncertain, retain the raw label and lower the confidence rather than guessing. Return JSON only. \par
\end{prompttablebox}

\begin{prompttablebox}{Prompt for Stage 1 Pass 2 visual description generation.}{tab:prompt_stage1_pass2}{Stage 1 --- Pass 2: Visual description generation}
Analyze this image and refine the visual grounding record for each candidate entity listed below. \par
Describe only attributes that are directly visible in this specific image. \par
\smallskip
\textbf{Image size.} \{processed\_width\}\(\times\)\{processed\_height\} pixels. \par
\smallskip
\textbf{Entity list.} \par
\{entities\_str\} \par
\smallskip
\textbf{Output format (JSON array).} \par
[ \par
\hspace*{2.0em}\{ \par
\hspace*{4.0em}"canonical\_name": "entity name from the input", \par
\hspace*{4.0em}"clearly\_visible": true/false, \par
\hspace*{4.0em}"visual\_attributes": ["directly visible attribute 1", "directly visible attribute 2"], \par
\hspace*{4.0em}"confidence\_adjustment": -1.0 to 1.0 \par
\hspace*{2.0em}\} \par
] \par
\smallskip
Do not infer identity-specific facts from world knowledge. If an entity is not clearly grounded in the image, set "clearly\_visible" to false and leave "visual\_attributes" empty. Return JSON only. \par
\end{prompttablebox}

\subsection{Prompts for Entity Knowledge Enrichment and Seed Selection}
\label{appendix:prompts_stage2}

This stage enriches each extracted entity with Wikipedia-derived textual information.

\begin{prompttablebox}{Prompt for Stage 2 Wikipedia-based node detail extraction.}{tab:prompt_stage2_node_detail}{Stage 2 --- Node detail extraction}
Extract key information about "\{entity\_name\}" from the following text. \par
\smallskip
Text content: \par
\{content\} \par
\smallskip
Please extract: \par
1. One-sentence description of what the entity is \par
2. Main type of entity (person/place/event/object/concept) \par
3. Key attributes of the entity (achievements, characteristics, etc.) \par
4. Most related entities to this entity \par
\smallskip
Return JSON: \par
\{ \par
\hspace*{2.0em}"description": "One-sentence description", \par
\hspace*{2.0em}"entity\_type": "person/place/event/object/concept", \par
\hspace*{2.0em}"properties": \{ \par
\hspace*{4.0em}"key1": "value1", \par
\hspace*{4.0em}"key2": "value2" \par
\hspace*{2.0em}\}, \par
\hspace*{2.0em}"related\_entities": ["entity1", "entity2", "entity3"] \par
\} \par
\smallskip
Return JSON only, no other content. \par
\end{prompttablebox}

\subsection{Prompts for Evidence Chain Construction}
\label{appendix:prompts_stage3}

This stage constructs evidence chains by classifying entity types, selecting candidate next-step entities, validating directed relations, filtering overly familiar entities, and detecting semantically equivalent adjacent nodes. Repeated-node rejection and verified shorter-path detection are deterministic graph checks applied after relation validation rather than additional LLM classifications.

\begin{prompttablebox}{Prompt for Stage 3 single entity type classification.}{tab:prompt_stage3_entity_type_single}{Stage 3 --- Single entity type classification}
Please strictly determine whether the following entity is a specific/concrete entity or an abstract/generic concept. \par
\smallskip
Entity name: \{entity\} \par
\smallskip
Strict Classification Rules: \par
- \textbf{Specific or concrete:} a named entity that is uniquely identifiable. \par
\hspace*{1.0em}- Specific persons: "Elon Musk", "Taylor Swift" (rather than "musician" or "billionaire") \par
\hspace*{1.0em}- Specific organizations: "Apple Inc.", "Harvard University" (rather than "company" or "university") \par
\hspace*{1.0em}- Specific locations: "Eiffel Tower", "Paris" (rather than "city" or "building") \par
\hspace*{1.0em}- Specific products: "iPhone 15", "Coca-Cola" (rather than "phone" or "soda") \par
\hspace*{1.0em}- Specific events: "2024 Olympics", "World War II" (rather than "war" or "sports event") \par
\hspace*{1.0em}- Specific artworks: "Mona Lisa", "Harry Potter" (rather than "book" or "painting") \par
\smallskip
- \textbf{Abstract or generic:} non-specific, generic, or conceptual terms. \par
\hspace*{1.0em}- Generic categories: "technology", "music", "food" \par
\hspace*{1.0em}- Emotions: "happiness", "love" \par
\hspace*{1.0em}- Concepts: "justice", "freedom", "democracy" \par
\hspace*{1.0em}- Generic roles: "teacher", "doctor" \par
\smallskip
Classification options: \par
- Person (\texttt{PER}): specific person name (e.g., "Elon Musk") \par
- Organization (\texttt{ORG}): specific company or institution (e.g., "Apple Inc.") \par
- Location (\texttt{LOC}): specific location (e.g., "Eiffel Tower") \par
- Product (\texttt{PROD}): specific product or brand (e.g., "iPhone 15") \par
- Event (\texttt{EVENT}): specific event (e.g., "2024 Olympics") \par
- Work of art (\texttt{WOA}): specific artwork (e.g., "Mona Lisa") \par
- Animal (\texttt{ANIMAL}): specific animal (e.g., "Golden Retriever") \par
- Plant (\texttt{PLANT}): specific plant (e.g., "Rose") \par
- Vehicle (\texttt{VEHICLE}): specific vehicle (e.g., "Boeing 747") \par
- Food (\texttt{FOOD}): food item (e.g., "Sushi", "Pizza") \par
- Dish (\texttt{DISH}): specific dish name (e.g., "Kung Pao Chicken") \par
- Material (\texttt{MATERIAL}): material (e.g., "Steel", "Cotton") \par
- Building (\texttt{BUILDING}): building (e.g., "Empire State Building") \par
- Brand (\texttt{BRAND}): brand name (e.g., "Nike") \par
- Technology (\texttt{TECH}): technology (e.g., "AI", "Blockchain") \par
- Medical (\texttt{MEDICAL}): medical term (e.g., "Diabetes") \par
- Natural phenomenon (\texttt{NAT}): e.g., "Earthquake" or "Rainbow" \par
- Other (\texttt{OTHER}): other specific or concrete entity \par
- Abstract (\texttt{ABSTRACT}): abstract concept or generic term \par
\smallskip
Output format (JSON): \par
\{"entity\_type": "ORG", "confidence": 0.95, "reasoning": "reason"\} \par
\smallskip
Return JSON only, no other content. \par
\end{prompttablebox}

\begin{prompttablebox}{Prompt for Stage 3 batch entity type classification.}{tab:prompt_stage3_entity_type_batch}{Stage 3 --- Batch entity type classification}
Classify the following \{n\} entities as specific/concrete or abstract/generic. \par
\smallskip
Entities to classify: \par
\{entities\_list\} \par
\smallskip
[Same classification rules and options as the single-entity prompt above.] \par
\smallskip
Output format (JSON array): \par
[ \par
\hspace*{2.0em}\{"entity": "Entity1", "entity\_type": "ORG", "confidence": 0.95, "reasoning": "reason"\} \par
] \par
\smallskip
Return JSON array only, no other content. \par
\end{prompttablebox}

\begin{prompttablebox}{Prompt for Stage 3 single-pair relation validation.}{tab:prompt_stage3_nli_single}{Stage 3 --- Single NLI relation validation}
Determine whether the proposed directed relation is factually supported by the supplied knowledge record. Use only the supplied evidence; do not validate a relation from model memory alone. \par
\smallskip
Source: \{source\_entity\} (QID: \{source\_qid\}) \par
Target: \{target\_entity\} (QID: \{target\_qid\}) \par
Original predicate: \{property\_id\} --- \{property\_label\} \par
Proposed direction and qualifiers: \{direction\_and\_qualifiers\} \par
Proposed normalized type: \{relation\_type\} \par
\smallskip
Supporting knowledge record: \par
\{knowledge\_record\} \par
Source URL and revision: \{source\_url\}, \{source\_revision\} \par
\smallskip
Valid types (mutually exclusive): \par
- part\_whole, member\_collection, causal \par
- temporal, spatial, comparative \par
- attributive, none \par
\smallskip
Validation rules: \par
- Confirm the original predicate, direction, and any qualifiers from the supplied record. \par
- Preserve relation polarity, such as before/after, inside/near, or larger/smaller, in relation\_detail. \par
- Return none when the evidence is absent, indirect, ambiguous, or inconsistent with the proposed edge. \par
- Set has\_relation to false if and only if relation\_type is none; otherwise set it to true. \par
\smallskip
Return JSON: \par
\{"has\_relation": true/false, "property\_id": "Pxxx", "raw\_relation": "predicate label", "relation\_type": "type", "relation\_detail": "directed or comparative predicate", "direction": "source\_to\_target/target\_to\_source/symmetric", "qualifiers": \{\}, "confidence": 0.0-1.0, "evidence\_span": "supporting span from the supplied record", "source\_url": "...", "source\_revision": "..."\} \par
\smallskip
Return JSON only. \par
\end{prompttablebox}

\begin{prompttablebox}{Prompt for Stage 3 batch relation validation.}{tab:prompt_stage3_nli_batch}{Stage 3 --- Batch NLI relation validation}
Validate every proposed directed relation in the following batch against its supplied knowledge record. Each record contains source and target names and QIDs, the original predicate and property ID, direction and qualifiers, an evidence span, source URL, and source revision. Use only these records and do not rely on model memory alone. \par
\smallskip
Relation records: \par
\{relation\_records\_json\} \par
\smallskip
\textbf{Types} (choose one; categories are mutually exclusive): \par
- part\_whole: A is part of B; preserve the direction explicitly \par
- member\_collection: A belongs to group B \par
- causal: A causes B \par
- temporal: A happens before/after/simultaneously with B; preserve the temporal direction \par
- spatial: A is located near/inside/far from B; preserve the spatial relation \par
- comparative: A is bigger/smaller/better/worse than B \par
- attributive: A has property B \par
- none: no relation \par
\smallskip
Preserve the polarity of temporal, spatial, and comparative predicates in relation\_detail. Return none when the supplied evidence does not directly support the proposed edge. Set has\_relation to false if and only if relation\_type is none. \par
\smallskip
Output format (JSON array): \par
[ \par
\hspace*{2.0em}\{"index": 1, "has\_relation": true, "property\_id": "Pxxx", "raw\_relation": "predicate label", "relation\_type": "part\_whole", "relation\_detail": "directed predicate", "direction": "source\_to\_target", "qualifiers": \{\}, "confidence": 0.9, "evidence\_span": "supporting span", "source\_url": "...", "source\_revision": "..."\} \par
] \par
\smallskip
Return JSON array only. \par
\end{prompttablebox}

\begin{prompttablebox}{Prompt for Stage 3 next-step entity selection.}{tab:prompt_stage3_next_hop}{Stage 3 --- Next-step entity selection}
Select the next-step entities for an evidence chain. \par
\smallskip
Current entity: \{current\_entity\} \par
\smallskip
Current entity properties (hints): \par
\{props\_text\} \par
\smallskip
Candidate neighbors (choose up to \{top\_k\}): \par
\{cand\_lines\} \par
\smallskip
Selection rules: \par
- Prefer candidates that are strongly suggested by the properties/hints. \par
- Prefer more obscure/less-known entities when ties (lower freq is more obscure). \par
- Avoid generic/abstract concepts. \par
- Choose a diverse set (do not pick near-duplicates). \par
\smallskip
Return JSON only: \par
\{"indices": [1, 2, 3]\} \par
\end{prompttablebox}

\begin{prompttablebox}{Prompt for Stage 3 entity familiarity assessment.}{tab:prompt_stage3_familiarity}{Stage 3 --- Entity familiarity assessment}
You are a knowledge graph expert. Assess how familiar/common each entity is to the general public. \par
\smallskip
Entities to assess: \par
\{entities\_list\} \par
\smallskip
For each entity, rate its familiarity on a scale of 0 to 1: \par
- 0.0 = extremely obscure/uncommon (e.g., a small village, niche technical term) \par
- 0.5 = moderately known (e.g., a well-known company, city, or historical event) \par
- 1.0 = extremely common/famous (e.g., a world-famous person, major country, common concept) \par
\smallskip
Return JSON only: \par
\{"familiarity\_scores": [\{"entity": "entity\_name", "score": 0.5\}, ...]\} \par
\smallskip
Notes: \par
- Focus on how familiar the entity is to an average educated person \par
- Judge name recognition independently for each entity; do not adjust a score to satisfy a downstream acceptance threshold \par
- Use the full scale and apply the same calibration across the batch \par
\end{prompttablebox}

\begin{prompttablebox}{Prompt for Stage 3 semantic duplicate detection.}{tab:prompt_stage3_duplicate}{Stage 3 --- Semantic duplicate detection}
You are a knowledge graph reasoning expert. Analyze whether directly connected nodes in the following evidence chain refer to the same real-world entity, including aliases, redirects, spelling variants, and genuine near-duplicate references. \par
\smallskip
Seed entity: \{seed\_entity\} \par
Current evidence chain nodes: \{entities\_in\_chain\} \par
\smallskip
Determine if there are duplicate nodes in the chain. If yes, provide the node pairs that need to be merged and the suggested merged entity name. \par
\smallskip
Output format (JSON): \par
\{ \par
\hspace*{2.0em}"has\_duplicates": true/false, \par
\hspace*{2.0em}"duplicate\_groups": [ \par
\hspace*{4.0em}\{"nodes": ["Node A", "Node B"], "merged\_name": "Merged Name"\} \par
\hspace*{2.0em}] \par
\} \par
\smallskip
Notes: \par
1. If no duplicates, return "has\_duplicates": false \par
2. Do not merge parent--child, part--whole, member--collection, or merely related entities \par
3. Use the canonical name of the shared real-world entity as the merged name \par
4. Only consider duplicate relationships between directly connected nodes \par
\end{prompttablebox}

% ==============================================================================
% Stage 4: Evidence-Grounded Query Construction
% ==============================================================================

\subsection{Prompts for Evidence-Grounded Query Construction}
\label{appendix:prompts_stage4}

This stage generates evidence-grounded textual queries from verified evidence chains. Visual task transformation and anti-leakage verification are performed in Stage~5.

\begin{prompttablebox}{Prompt for Stage 4 long-horizon query generation.}{tab:prompt_stage4_question_generation}{Stage 4 --- Long-horizon query generation}
Based on the following evidence chain, generate a long-horizon evidence-seeking query. \par
\smallskip
Evidence Chain: \par
\{reasoning\_chain\_desc\} \par
\smallskip
\{start\_info\} \par
\smallskip
\{end\_info\} \par
\smallskip
1. Evidence-chain fidelity: \par
- Construct the query from the visual seed toward the target so that every required clue corresponds to a verified relation in the supplied chain. \par
\smallskip
2. Obfuscation \& Vagueness: \par
- Never directly mention the target or any intermediate entities by name. \par
- Describe entities indirectly using attributes, categories, or general characteristics. \par
- Avoid giving away the reasoning path; each step must be subtle. \par
\smallskip
3. Difficulty \& Multi-Step Search: \par
- The question should require exactly $k$ evidence steps, where $k$ is the number of relational edges in the chain. \par
- Each reasoning step should provide a vague constraint or clue that requires external search or knowledge verification. \par
\smallskip
4. Clarity of Requirements for Generation: \par
- Maintain coherence and fluency. \par
- Keep the verified reasoning chain intact. \par
- Ensure the question is solvable via logical reasoning over external information, without revealing shortcuts. \par
\smallskip
\textbf{Anti-leakage requirements.} \par
1. Never directly mention the target in the question text \par
2. Make clues appropriately indirect by using descriptions rather than specific names \par
3. Describe entities by their attributes rather than their names \par
4. Avoid giving away the reasoning path \par
5. The question should read as a natural information-seeking query that requires multi-step search and reasoning \par
6. Use relative or comparative descriptions instead of direct references \par
\hspace*{1.5em}- Avoid: "Which company founded by Elon Musk..." \par
\hspace*{1.5em}- Prefer: "Which aerospace company founded..." \par
\smallskip
Please generate a natural question whose target is the last entity in the chain: \par
\smallskip
\textasciigrave{}\textasciigrave{}\textasciigrave{}json \par
\{ \par
\hspace*{2.0em}"question": "Generated question", \par
\hspace*{2.0em}"constraints": ["constraint1", "constraint2", ...], \par
\hspace*{2.0em}"reasoning\_path": ["reasoning step 1", "reasoning step 2", "..."] \par
\} \par
\textasciigrave{}\textasciigrave{}\textasciigrave{} \par
\smallskip
Return JSON only, no other content. \par
\end{prompttablebox}

% ==============================================================================
% Stage 5: Visual DeepSearch Task Generation and Anti-Leakage Verification
% ==============================================================================

\subsection{Prompts for Visual DeepSearch Task Generation and Anti-Leakage Verification}
\label{appendix:prompts_stage5}

This stage first replaces the root entity with a visual reference and simplifies the resulting task query. It then applies the \textsc{Solver}--\textsc{Judge}--\textsc{Rewrite} loop to the final user-visible query for at most $R_{\max}=3$ rounds, followed by clue-necessity verification. Text-only leakage detection and clue necessity are treated as separate checks: the former identifies textual shortcuts, whereas the latter combines a deterministic evidence-graph test with a paired, blind image-and-search Solver test to determine whether every annotated clue is required by the intended path. Unresolved queries are discarded. The transformation and simplification prompts are listed later in this subsection, but they are executed before the anti-leakage prompts.

\begin{prompttablebox}{Prompt for Stage 5 text-only Solver agent.}{tab:prompt_stage5_solver}{Stage 5 --- Text-only Solver Agent}
You are a Solver Agent. \par
\smallskip
Task: \par
Given only the final task-query text, try to recover the target without access to the image, image crop, or annotated evidence chain. \par
Focus on whether the query text alone reveals enough information to determine the target. \par
\smallskip
Instructions: \par
- Read the question carefully. \par
- If you can determine the target from the query text alone, provide your response. \par
\smallskip
Input (JSON): \par
\{batch\_json\} \par
\smallskip
Output format (JSON only): \par
\{ \par
\hspace*{1.0em}"solves": [ \par
\hspace*{2.0em}\{ \par
\hspace*{3.0em}"original\_index": 1, \par
\hspace*{3.0em}"solver\_response": "...", \par
\hspace*{3.0em}"solver\_confidence": 0.0-1.0, \par
\hspace*{3.0em}"used\_text\_clues": "What specific text clues did you use to determine the target?" \par
\hspace*{2.0em}\} \par
\hspace*{1.0em}] \par
\} \par
\smallskip
Return JSON only. Do not output any other text. \par
\end{prompttablebox}

\begin{prompttablebox}{Prompt for Stage 5 Judge agent.}{tab:prompt_stage5_judge}{Stage 5 --- Judge Agent}
You are a Judge Agent. \par
\smallskip
Task: \par
Evaluate every question in the input list and output one judgment for each item. \par
Do not skip any item. The input contains exactly \{len(batch)\} questions. \par
\smallskip
Input: \par
The input is a JSON list of questions, where each question contains: \par
- index: The question number \par
- question: The question text \par
- target: The annotated target \par
- solver\_response: The response given by the Solver \par
- solver\_confidence: The confidence score from Solver (0.0-1.0) \par
- used\_text\_clues: What clues the Solver used \par
\smallskip
\textbf{Evaluation steps} (repeat for each question): \par
\smallskip
Step 1: Check if the Solver response matches the target \par
- Compare solver\_response with target \par
- If they match -\textgreater{} correct; if they don't match or solver said "cannot determine" -\textgreater{} incorrect \par
\smallskip
Step 2: Check for information leakage or a text-only shortcut \par
- If the Solver is correct without the image, mark leakage\_detected: true, regardless of whether it used all or only part of the textual clues \par
- If the Solver is incorrect, independently inspect whether the target name, a unique alias, or a near-deterministic textual clue is exposed \par
- Record which textual clues create the shortcut \par
\smallskip
Step 3: Check task validity \par
- Determine whether the query is ambiguous, under-specified, non-unique, or inconsistent with the intended evidence chain \par
\smallskip
Output Requirements: \par
- Output a judgment for every question in the input \par
- Use the Markdown format below \par
\smallskip
Output Format (Markdown): \par
\smallskip
\#\# Judgments \par
\smallskip
\#\#\# original\_index=0, solve\_result=correct/incorrect \par
- leakage\_detected: true/false \par
- leakage\_type: none/direct\_target/alias/redundant\_clue/text\_only\_shortcut \par
- ambiguity\_detected: true/false \par
- uniqueness\_issue: true/false \par
- leakage\_reason: ... \par
- judge\_feedback: ... \par
\smallskip
[Repeat for each question.] \par
\smallskip
Use the original\_index from the input (e.g., 0, 1, 2...) as the section header. \par
\smallskip
Return the judgments in Markdown format as shown above. \par
\end{prompttablebox}

\begin{prompttablebox}{Prompt for Stage 5 question rewriting.}{tab:prompt_stage5_rewrite}{Stage 5 --- Question rewriting prompt}
You are a Generator Agent. \par
\smallskip
Goal: \par
Rewrite each task query based on the Judge's feedback to eliminate information leakage while maintaining the long-horizon evidence structure. \par
The question must remain solvable from the image-grounded evidence chain, but the target should not be inferable from the question text alone. \par
\smallskip
Evidence Chain Based Question Generation Rules: \par
1. Evidence-chain fidelity: Preserve every relation required by the verified chain. \par
2. Obfuscation \& Vagueness: Never directly mention the final target or intermediate entities. \par
3. Difficulty \& Multi-Step Search: Each reasoning step provides an indirect but verifiable constraint. \par
4. Anti-Leakage Requirements: Use indirect descriptions and no identity-revealing names. \par
\smallskip
Instructions: \par
- Use the "judge\_feedback" field to identify why the question leaks and how to fix it. \par
- The visual entity in the root node should be replaced with a visual reference. \par
- Do not reveal the identity of the visual entity through text descriptions. \par
- Keep the reasoning chain (intermediate nodes and relationships) that leads to the target. \par
- Use vague/indirect descriptions instead of specific names for entities. \par
- Keep the language consistent with the original English. \par
\smallskip
\textbf{Input format (JSON).} \par
\{batch\_json\} \par
\smallskip
\textbf{Output format (JSON only).} \par
\{ \par
\hspace*{1.0em}"rewrites": [ \par
\hspace*{2.0em}\{ \par
\hspace*{3.0em}"original\_index": 1, \par
\hspace*{3.0em}"rewritten\_question": "..." \par
\hspace*{2.0em}\} \par
\hspace*{1.0em}] \par
\} \par
\smallskip
Return JSON only. Do not output any other text. \par
\end{prompttablebox}

% ==============================================================================
% Stage 5: Visual task transformation and simplification prompts
% ==============================================================================

\paragraph{Visual task transformation and simplification prompts.}

The following prompts implement the transformation and simplification operations executed at the beginning of Stage~5. They preserve visual grounding while removing textual clues that could reveal the visual entity.

\begin{prompttablebox}{Prompt for Stage 5 root-entity grounding and Visual DeepSearch task transformation.}{tab:prompt_stage5_visual_task_transform}{Stage 5 --- Root-entity grounding and Visual DeepSearch task transformation}
You are an expert at converting evidence-grounded queries into Visual DeepSearch tasks. \par
\smallskip
Your task is to: \par
1. Use the supplied root entity and its verified image-grounding record; do not infer the root entity from the query text \par
2. Replace the root-entity name with a natural, non-identifying visual reference \par
3. Preserve the verified evidence chain and its target \par
\smallskip
\textbf{Input fields.} \par
- question: the evidence-grounded textual query \par
- target: the annotated target \par
- root\_entity: the verified seed entity \par
- root\_entity\_type: its entity type \par
- root\_bbox: its verified bounding box \par
- root\_visual\_attributes: directly visible, non-identifying attributes \par
\smallskip
\textbf{Transformation rules.} \par
- Replace only the root-entity mention; do not alter the target or relations required by the chain \par
- Select a natural visual reference based on root\_entity\_type, for example: \par
\hspace*{1.0em}- Person: "this person", "the individual", "the man/woman" \par
\hspace*{1.0em}- Place: "this location", "this place", "this spot", "the area" \par
\hspace*{1.0em}- Object: "this object", "this item", "the thing", "the structure" \par
\hspace*{1.0em}- Organization: "this organization/company", "the group" \par
\hspace*{1.0em}- Event: "this event", "the occasion" \par
- Refer explicitly to the supplied image using natural wording such as "shown in the image" or "visible in the image" \par
- Do not insert a visual attribute if it reveals the root identity or makes the task solvable from text alone \par
- If multiple same-type instances are present, use root\_bbox and non-identifying spatial attributes to form a unique reference (e.g., "the leftmost vehicle in the image") \par
- If a unique non-identifying reference cannot be formed, mark the item invalid instead of using an ambiguous phrase such as "one of the vehicles" \par
\smallskip
Input: \par
\{questions\_json\} \par
\smallskip
\textbf{Output format (JSON).} \par
\{ \par
\hspace*{2.0em}"transformed\_questions": [ \par
\hspace*{4.0em}\{ \par
\hspace*{6.0em}"original\_index": 1, \par
\hspace*{6.0em}"root\_entity": "verified root entity", \par
\hspace*{6.0em}"target": "annotated target", \par
\hspace*{6.0em}"task\_query": "transformed query with visual reference", \par
\hspace*{6.0em}"visual\_reference": "non-identifying reference to the grounded region", \par
\hspace*{6.0em}"valid\_visual\_reference": true/false, \par
\hspace*{6.0em}"invalid\_reason": "empty when valid" \par
\hspace*{4.0em}\} \par
\hspace*{2.0em}] \par
\} \par
\smallskip
Return JSON only, no other content. \par
\end{prompttablebox}

\begin{prompttablebox}{Prompt for Stage 5 question simplification.}{tab:prompt_stage5_simplification}{Stage 5 --- Question simplification}
\{questions\_json\} \par
\smallskip
You are an expert at rewriting Visual DeepSearch task queries so that their targets can be recovered only by inspecting the image, not from the query text alone. \par
\smallskip
\textbf{Rule 1: Remove identity-revealing descriptions of the visual entity.} \par
The query may contain descriptive phrases that identify the visual entity through text alone. Remove those phrases and replace them with a natural image reference. \par
- Preserve only non-identifying localization needed to select the verified region, such as "leftmost" or "in the foreground" \par
- Do not retain names, logos transcribed into text, unique aliases, or descriptions that independently determine the root identity \par
- Keep an explicit reference to the supplied image, but do not require one fixed surface phrase \par
\smallskip
\textbf{Rule 2: Use plain language.} \par
Replace verbose/pompous vocabulary with simple equivalents: \par
- "sovereign jurisdiction" -\textgreater{} "country" \par
- "fiscal barometer / financial benchmark" -\textgreater{} "stock index" \par
- "headquartered / administrative home" -\textgreater{} "based in" \par
- "architect / pioneer of" -\textgreater{} "maker of" / "company behind" \par
\smallskip
\textbf{Rule 3: Remove unnecessary modifiers.} \par
Delete decorative adjectives or adverbs that do not affect the target: \par
- Delete terms such as "renowned", "prominent", "leading", "prestigious", and "iconic" \par
\smallskip
\textbf{Rule 4: Generalize nonessential numbers and dates.} \par
When exact numbers or dates are not essential for determining the target, delete or generalize them: \par
- Delete them if they are not crucial, such as "30", "500", "1990", or "mid-2010s" \par
- Keep them only when crucial for determining the target (e.g., "top 3", "first", "largest") \par
\smallskip
\textbf{Rule 5: Preserve the reasoning chain.} \par
Keep all facts that are part of the logical path to the target: \par
- Named entities that do not describe the visual entity (e.g., index names, company names, place names) \par
- Quantitative facts only when crucial for reasoning \par
- Unique relationships: "subsidiary of", "built by", "world's largest" \par
\smallskip
\textbf{Output format (JSON).} \par
\{ \par
\hspace*{2.0em}"simplified\_questions": [ \par
\hspace*{4.0em}\{ \par
\hspace*{6.0em}"original\_index": 1, \par
\hspace*{6.0em}"original\_question": "...", \par
\hspace*{6.0em}"simplified\_question": "rewritten question" \par
\hspace*{4.0em}\} \par
\hspace*{2.0em}] \par
\} \par
\smallskip
Return JSON only, no other content. \par
\end{prompttablebox}

\begin{prompttablebox}{Prompt for Stage 5 clue-necessity verification.}{tab:prompt_stage5_clue_necessity}{Stage 5 --- Clue-necessity verification}
Run this check after the Solver--Judge--Rewrite loop has converged. This is a necessity test, not a leakage-attribution test. \par
\smallskip
For each final task query, map every atomic evidence-bearing clue to its verified edge or constraint and create one ablated query by removing exactly that clue. A deterministic graph routine recomputes reachability, alternative paths, target uniqueness, and the evidence-step count after each ablation. Separately, an image-and-search-enabled Solver receives the image and either the original or ablated query under the same tool budget, but never receives the target, accepted aliases, evidence graph, or annotated path. \par
\smallskip
\textbf{Inputs.} \par
- Queries, clue-to-edge mappings, targets and accepted aliases, verified evidence graphs, and evidence-step counts: \{ablation\_batch\_json\} \par
- Deterministic graph-check records for the original and ablated constraints: \{graph\_necessity\_json\} \par
- Blind image-and-search Solver outputs for paired original and ablated queries: \{paired\_solver\_results\_json\} \par
- Text-only Solver--Judge records rerun after any shortening: \{text\_only\_recheck\_json\} \par
\smallskip
\textbf{Requirements.} \par
- Ablate exactly one clue at a time; keep all other wording unchanged except for minimal grammatical repair \par
- Do not add new clues during grammatical repair \par
- Mark a clue redundant if the remaining graph constraints still identify a unique target, a verified alternative path remains, or the paired Solver recovers the target from the ablated query \par
- Use paired Solver evidence only when the same Solver recovers the target from the original query; Solver failure on an ablated query alone is not evidence of necessity \par
- Mark a clue necessary only when removing it eliminates unique target reachability in the verified evidence graph and no valid paired recovery is observed \par
- For a redundant clue, shorten the task, recompute the reasoning path and evidence-step count, and rerun the text-only shortcut check; discard the task if it no longer satisfies the construction constraints or contains a shortcut \par
- Retain a task only if every annotated clue is necessary; send uncertain cases for human review \par
\smallskip
\textbf{Output format (JSON only).} \par
\{ \par
\hspace*{2.0em}"necessity\_checks": [ \par
\hspace*{4.0em}\{ \par
\hspace*{6.0em}"original\_index": 1, \par
\hspace*{6.0em}"removed\_clue": "...", \par
\hspace*{6.0em}"ablated\_query": "...", \par
\hspace*{6.0em}"unique\_target\_reachable\_without\_clue": true/false, \par
\hspace*{6.0em}"verified\_alternative\_path": true/false, \par
\hspace*{6.0em}"original\_query\_solved": true/false, \par
\hspace*{6.0em}"ablated\_query\_solved": true/false, \par
\hspace*{6.0em}"text\_only\_shortcut\_after\_shortening": true/false/null, \par
\hspace*{6.0em}"clue\_status": "necessary/redundant/uncertain", \par
\hspace*{6.0em}"recomputed\_evidence\_steps": 0, \par
\hspace*{6.0em}"decision": "retain/shorten/discard/human\_review" \par
\hspace*{4.0em}\} \par
\hspace*{2.0em}] \par
\} \par
\smallskip
Return JSON only, no other content. \par
\end{prompttablebox}

% ==============================================================================
% Stage 6: Multi-Chain Fusion
% ==============================================================================

\subsection{Prompts for Multi-Chain Fusion}
\label{appendix:prompts_stage6}

This stage constructs fused tasks from multiple Visual DeepSearch task items. 
The prompts extract verifiable intermediate values from targets or chain metadata and select a numerical or conditional fusion operation. A deterministic program then validates the normalized inputs and computes the final target before an LLM generates a unified task query. Component-necessity records are generated by applying the Stage~5 clue-necessity protocol after removing each complete component clue. Anchor-necessity records map every component to at least one grounded region and contain paired component-target log-likelihoods for the original image and an image with only that region masked, computed with the fixed visual-dependence scorer and the operational threshold described in Section~\ref{sec:benchmark_quality}. Every newly generated fused query subsequently undergoes text-only shortcut, target-uniqueness, component-necessity, visual-anchor-necessity, and deterministic-result verification.

\begin{prompttablebox}{Prompt for Stage 6 intermediate-value extraction.}{tab:prompt_stage6_intermediate_value}{Stage 6 --- Intermediate-value extraction}
Extract an existing, verifiable intermediate value from the task target or its evidence-chain metadata. \par
\smallskip
Question: \{context\} \par
Target: "\{target\}" \par
Chain metadata and supporting evidence: \par
\{chain\_metadata\} \par
\smallskip
Allowed value types include: \par
- an explicit year or date \par
- a count or ranking \par
- a numerical quantity with its unit \par
- a categorical or Boolean value used by an evidence-grounded conditional-selection rule \par
\smallskip
\textbf{Validity rules.} \par
- The value must be explicitly supported by the target or the supplied chain metadata and evidence \par
- Record the exact source field and supporting evidence used for extraction \par
- Preserve units and temporal reference points \par
- Do not encode strings using ASCII values, alphabet positions, character counts, word counts, vowel counts, or similar artificial proxies \par
- Do not invent a value or derive a time difference from the current year unless the reference date is fixed in the supplied metadata \par
- If no natural, unambiguous value can be extracted, return "valid": false; do not force a conversion \par
\smallskip
Output JSON only: \par
\{ \par
\hspace*{2.0em}"valid": true/false, \par
\hspace*{2.0em}"value": "extracted value", \par
\hspace*{2.0em}"value\_type": "year/date/count/ranking/quantity/categorical/boolean", \par
\hspace*{2.0em}"unit": "unit or null", \par
\hspace*{2.0em}"source\_field": "target or metadata field", \par
\hspace*{2.0em}"supporting\_evidence": "verifiable evidence", \par
\hspace*{2.0em}"reason": "why this extraction is valid or invalid" \par
\} \par
\smallskip
Return complete JSON only, with no additional text. \par
\end{prompttablebox}

\begin{prompttablebox}{Prompt for Stage 6 fusion operation decision.}{tab:prompt_stage6_operation}{Stage 6 --- Fusion operation decision}
Given $K\geq 2$ verified intermediate values from related task chains, select a semantically justified fusion operation. \par
\smallskip
Extractions: \par
\{extraction\_json\} \par
\smallskip
Available fusion operations: \par
1. Add (\texttt{ADD}): sum compatible numerical values \par
2. Subtract (\texttt{SUBTRACT}): subtract the second value from the first, for exactly two ordered values \par
3. Multiply (\texttt{MULTIPLY}): multiply compatible numerical values \par
4. Divide (\texttt{DIVIDE}): divide the first value by the nonzero second value, for exactly two ordered values \par
5. Maximum (\texttt{MAX}): select the item associated with the maximum compatible value \par
6. Minimum (\texttt{MIN}): select the item associated with the minimum compatible value \par
7. Average (\texttt{AVG}): compute the average of compatible numerical values \par
8. Conditional selection (\texttt{CONDITIONAL\_SELECT}): select a target or value using an explicit condition supported by the component chains \par
\smallskip
\textbf{Validity rules.} \par
- Use only inputs marked valid and preserve their order where the operation is non-commutative \par
- Numerical operations require compatible value types and units \par
- Reject division by zero, ambiguous extraction, incompatible units, and operations without a meaningful interpretation \par
- For average or division, specify an exact rounding rule only when needed \par
- Conditional selection must state the evidence-grounded condition and the possible selections \par
- If no operation is valid and meaningful, return "valid": false instead of forcing a fusion \par
\smallskip
Do not calculate the final target in this step. A deterministic routine applies the selected operation to the validated, normalized inputs after this decision. \par
\smallskip
Output JSON only: \par
\{ \par
\hspace*{2.0em}"valid": true/false, \par
\hspace*{2.0em}"operation": "ADD/SUBTRACT/MULTIPLY/DIVIDE/MAX/MIN/AVG/CONDITIONAL\_SELECT", \par
\hspace*{2.0em}"ordered\_inputs": ["component id 1", "component id 2"], \par
\hspace*{2.0em}"condition": "explicit condition or null", \par
\hspace*{2.0em}"rounding\_rule": "rule or null", \par
\hspace*{2.0em}"reason": "semantic justification or invalidity reason" \par
\} \par
\smallskip
Return JSON only, no other text. \par
\end{prompttablebox}

\begin{prompttablebox}{Prompt for Stage 6 fused question generation.}{tab:prompt_stage6_question_generation}{Stage 6 --- Fused question generation}
\{question\_section\} \par
\smallskip
Verified deterministic calculation record: \par
\{deterministic\_calculation\_json\} \par
\smallskip
Task: \par
Create a single, coherent question that: \par
1. Naturally integrates content from all \{n\} original Visual DeepSearch task queries about the image \par
2. Requires the solver to recover every component value through its original image-grounded evidence chain \par
3. States the verified \{operation\_display\} rule needed to derive the deterministically computed final target \par
4. Does not reveal component targets, intermediate values, or the final target \par
\smallskip
Intermediate-value guidance (for the reasoning record only): \par
\{extraction\_hints\} \par
- Final step: Apply the \{fusion\_operation\} operation to obtain the final target. \par
\smallskip
\textbf{Output format (JSON only).} \par
\{ \par
\hspace*{2.0em}"question": "A single natural question grounded in the supplied image", \par
\hspace*{2.0em}"final\_target": "exact value copied from deterministic\_calculation\_json", \par
\hspace*{2.0em}"reasoning\_path": ["component-chain step 1", "component-chain step 2", "fusion step"], \par
\hspace*{2.0em}"component\_ids": ["id1", "id2"] \par
\} \par
\smallskip
\textbf{Hard constraints.} \par
The "question" field must: \par
- Preserve the visual references and evidence-bearing clues of every selected component \par
- Not expose any component target, intermediate value, or final target \par
- Include a number or condition only when it is an indispensable part of the intended reasoning, never as an exposed intermediate result \par
- State the numerical or conditional fusion rule unambiguously \par
- Maintain visual grounding by explicitly referring to the supplied image \par
- Use natural, fluent language without robotic listing \par
The "final\_target" field must copy the verified result exactly; do not recalculate or alter it. \par
\smallskip
Return only the complete JSON object, without explanation or additional text. \par
\end{prompttablebox}

\begin{prompttablebox}{Prompt for Stage 6 post-fusion verification.}{tab:prompt_stage6_post_verification}{Stage 6 --- Post-fusion verification}
Validate a newly generated fused Visual DeepSearch item using the supplied verification records. Do not infer missing evidence or repair the item during validation. \par
\smallskip
\textbf{Inputs.} \par
- Fused query and proposed final target: \{fused\_item\_json\} \par
- Deterministic calculation record, including normalized inputs, units, operation, ordering, rounding rule, and verified result: \{deterministic\_calculation\_json\} \par
- Component queries, evidence chains, targets, and accepted aliases: \{component\_records\_json\} \par
- Blind text-only Solver and Judge results for the fused query: \{text\_only\_verification\_json\} \par
- Component-clue ablation records from the Stage 5 structural and paired-Solver protocol: \{component\_necessity\_json\} \par
- Per-anchor records containing component IDs, grounded regions, original and masked-image component-target log-likelihoods, score gaps, the 0.182 operational threshold, and any adjudicated review decisions: \{anchor\_necessity\_json\} \par
\smallskip
\textbf{Checks.} \par
1. Text-only shortcut: the fused query must pass the same Solver--Judge anti-leakage protocol as a single-chain query. \par
2. Target uniqueness: every component must have one unique normalized target, and the deterministic fusion of those targets must be single-valued. \par
3. Component necessity: each component-ablation record must show loss of unique reachability with no verified alternative path or valid paired-Solver recovery. \par
4. Visual-anchor necessity: every required component must have at least one grounded anchor, every recorded anchor must map to a required component, and each anchor--component score gap must exceed 0.182. A lower gap cannot pass automatically and requires an adjudicated masking review; missing coverage or failed review invalidates the item. \par
5. Deterministic result: the stored final target must exactly match the verified calculation record, including units and rounding. \par
\smallskip
Mark the item valid only if all five checks pass. Otherwise return rewrite or discard together with the failed checks. Any rewritten item must be reverified from the beginning. \par
\smallskip
\textbf{Output format (JSON only).} \par
\{ \par
\hspace*{2.0em}"valid": true/false, \par
\hspace*{2.0em}"text\_only\_shortcut\_free": true/false, \par
\hspace*{2.0em}"target\_unique": true/false, \par
\hspace*{2.0em}"all\_components\_necessary": true/false, \par
\hspace*{2.0em}"all\_components\_anchor\_covered": true/false, \par
\hspace*{2.0em}"all\_visual\_anchors\_necessary": true/false, \par
\hspace*{2.0em}"deterministic\_result\_verified": true/false, \par
\hspace*{2.0em}"failed\_checks": ["..."], \par
\hspace*{2.0em}"decision": "retain/rewrite/discard" \par
\} \par
\smallskip
Return JSON only, no other content. \par
\end{prompttablebox}
\section{Human Baseline Protocol and Aggregate Results}
\label{appendix:human_baseline_protocol}

\paragraph{Protocol.}
The three participants completed all 600 tasks independently and were not
allowed to communicate about individual items. They received only the query and
input image and had no access to the reference answer, annotated evidence chain,
construction metadata, or responses from the evaluated models. They could use
the same three classes of evidence tools available in the \emph{Search+Crop}
setting: Web search, reverse-image search, and image-region cropping.

\paragraph{Aggregation.}
Accuracy is first computed for each participant and then macro-averaged across
the three; it is not obtained by majority voting or collaborative adjudication.
Because all three participants complete the same number of items, the macro-average
is identical to the accuracy computed over all 1,800 individual judgments.
Table~\ref{tab:human_baseline_details} reports the number of tasks and judgments
in each difficulty subset together with macro-averaged Pass@1. The overall
result is the task-count-weighted mean of the L2 and L3 results using the
benchmark composition. We further report the distribution of incorrect
responses under the predefined taxonomy.

\begin{table}[htbp]
\centering
\scriptsize
\caption{Human performance by difficulty level.}
\label{tab:human_baseline_details}
\setlength{\tabcolsep}{3.5pt}
\begin{tabular}{lrrr}
\toprule[1.2pt]
\textbf{Subset} & \textbf{\# Tasks} & \textbf{\# Judgments} & \textbf{P@1} \\
\midrule
L2 & 154 & 462 & 81.17 \\
L3 & 446 & 1,338 & 78.92 \\
Overall & 600 & 1,800 & 79.50 \\
\bottomrule[1.2pt]
\end{tabular}
\vspace{-6pt}
\end{table}

\paragraph{Error analysis.}
We categorized incorrect judgments using a pre-specified taxonomy informed
by the human-search protocols of BrowseComp~\cite{wei2025browsecomps} and
BrowseComp-VL~\cite{geng2026webwatcher} and by our model-trajectory taxonomy.

\begin{table}[htbp]
\centering
\scriptsize
\caption{Distribution of incorrect human judgments by failure type.}
\label{tab:human_error_analysis}
\setlength{\tabcolsep}{3pt}
\begin{tabularx}{\linewidth}{X r}
\toprule[1.2pt]
\textbf{Failure type} & \textbf{Share} \\
\midrule
Incorrect or missing visual grounding & 14.9\% \\
Retrieval or evidence-traversal error & 72.8\% \\
Fusion, calculation, or normalization error & 12.3\% \\
\midrule
Total & 100.0\% \\
\bottomrule[1.2pt]
\end{tabularx}
\vspace{-6pt}
\end{table}

\section{Ethics, Licensing, and Responsible Use}
\label{appendix:ethics_licensing_responsible_use}

\paragraph{Image provenance, licensing, and redistribution.}
We distinguish public accessibility from permission to redistribute. The
project maintains an item-level provenance ledger containing the source page,
creator or rights holder when available, license name and version, license URL,
required credit, and local modification status. The license attached to our
code and annotations does not relicense third-party images. An image is
eligible for redistribution only when its recorded terms permit redistribution
and the associated attribution and share-alike obligations can be preserved.
For material whose status is unclear or whose terms do not permit
redistribution, a release should contain only the source URL and benchmark
metadata rather than a copy of the image. License status is a snapshot at the
time of collection and should be rechecked before downstream redistribution.

\paragraph{People, personal information, removal, and appeal.}
Some tasks contain recognizable people, public events, storefronts, logos, or
other real-world scenes. We do not add biometric templates, private contact
details, or other intentionally collected sensitive personal data. Nevertheless,
some tasks may support identity or location inference from public visual and Web
evidence, which creates privacy and contextual-integrity risks even when an
image is lawfully accessible. Users should not apply \bench to identify private
individuals, track people, infer sensitive traits, or make decisions affecting
employment, education, insurance, credit, policing, or access to services.
Rights holders and depicted individuals may request correction, restricted
redistribution, or removal through the corresponding-author contact listed in
the paper. Substantiated requests will be recorded in the dataset change log;
affected image bytes will be removed from the next release, and the associated
task will be withdrawn or replaced. A requester who disagrees with the initial
decision may ask for review by a second maintainer who was not responsible for
the first decision.

\paragraph{Coverage and representational limitations.}
\bench is an English-language benchmark assembled from sources that are
available and searchable on the open Web. Its five broad categories and 25
scenarios do not constitute a representative sample of the world's regions,
languages, cultures, people, or entity types. Source availability, licensing,
search-engine ranking, and annotator expertise can overrepresent well-documented
entities, English-language pages, globally prominent people and institutions,
and regions with extensive digitized cultural material. They can
underrepresent low-resource languages, less-connected communities, private or
informal settings, and culturally specific interpretations. Performance should
therefore not be interpreted as geographic, linguistic, or cultural parity, and
the benchmark should not be used to rank the importance or ``searchability'' of
people or cultures. Future versions should report geographic, language,
cultural, and entity-type distributions and use those audits to guide targeted
collection rather than treating aggregate accuracy as a fairness measure.

\paragraph{Potential misuse and intended scope.}
The benchmark is intended for research on multimodal retrieval, visual
grounding, evidence tracing, and robustness. The same capabilities could be
misused for automated surveillance, doxxing, unwanted identity or location
inference, large-scale profiling, copyright-violating media collection, or the
generation of plausible but unsupported claims about real people. We recommend
human review for any real-world use, provenance-preserving outputs, rate limits
for identity- or location-oriented queries, and refusal or escalation policies
for requests involving private persons or sensitive attributes. Benchmark
scores are not evidence that a system is safe for deployment.

\section{Dynamic Web Retrieval and Temporal Robustness}
\label{appendix:retrieval_reproducibility}

Live Web search inevitably varies across time, locations, and search providers.
Such variation affects the evidence available to an agent, but not the
definition of our benchmark tasks. When an answer may change over time, the
query includes an explicit temporal anchor, ensuring that later updates do not
alter the intended target. Changes in rankings, page availability, or newly
indexed content may introduce additional retrieval noise and increase search
difficulty, rather than invalidate the benchmark. For reproducible evaluation,
we separate fixed task annotations from dynamic retrieval results and record
the retrieval provider, timestamp, queries, ranked results, and caching
protocol.

\section{Reverse-Image Search and Non-Indexed Image Diagnostic}
\label{appendix:nonindexed_images}
Reverse-image search is used only to assist visual identification, rather than
to retrieve the final answer directly. Its returned titles and thumbnails may
reveal the source image, depicted entities, or scene context, but they do not by
themselves resolve the query. Since the target is separated from the visual
clues by multi-step evidence chains and, for fused tasks, an additional
composition step, the agent must still retrieve and verify the intermediate
evidence to derive the final answer.

\paragraph{Retrieval backends and evaluation window.}
\textit{Text Search} uses Google Search results accessed through the Serper Search
API (\texttt{https://google.serper.dev/search}). The hosted search service does
not expose a fixed ranking-model version; we therefore record the provider,
query, returned rank, URL, title, snippet, and retrieval timestamp. All results
reported in the paper were collected in 2026.
The search depth is fixed to the top three organic results. For each result,
\eval attempts to fetch the linked page, truncates the extracted text to 30,000
characters, and summarizes it with the fixed Qwen3-32B summarizer described in
Section~\ref{sec:experimental_setup}.

\section{Evaluation and Inference Prompts}
\label{appendix:evaluation_prompts}

This section reports the prompts used by \eval at inference and scoring time.
In the tool-use setting, the dataset-specific base system prompt is concatenated
with the tool-use prompt below. The task image and query are then supplied in the
user message. Text enclosed in braces denotes a runtime placeholder. The tool
schemas are rendered as JSON by the evaluation program before being inserted at
\{tool\_definitions\}. We use the same tool names as in the main text:
\textit{Text Search}, \textit{Image Search}, and \textit{Image Crop}. For exact
reproducibility, each tool's code-level function identifier is also reported.

\subsection{Tool-Use Prompt and Tool Definitions}

\begin{prompttablebox}{System-prompt suffix defining the tool-call protocol.}{tab:prompt_eval_tools}{Evaluation --- Tool-use system-prompt suffix}
\# Tools \par
\smallskip
You may call at most one function per response/round to assist with the user query. \par
\smallskip
You are provided with function signatures within $<$tools$>$$<$/tools$>$ XML tags: \par
$<$tools$>$ \par
\{tool\_definitions\} \par
$<$/tools$>$ \par
\smallskip
For each function call, return a json object with function name and arguments within $<$tool\_call$>$$<$/tool\_call$>$ XML tags: \par
$<$tool\_call$>$ \par
\{"name": $<$function-name$>$, "arguments": $<$args-json-object$>$\} \par
$<$/tool\_call$>$ \par
\end{prompttablebox}

\begin{prompttablebox}{Tool schema for image-region inspection.}{tab:prompt_eval_crop_tool}{Evaluation --- Image Crop}
Name: Image Crop \par
Implementation identifier: image\_zoom\_in\_tool \par
Description: Zoom in on a specific region of an image by cropping it based on a bounding box (bbox) and an optional object label. \par
\smallskip
Arguments: \par
- bbox\_2d (required array of four numbers): The bounding box of the region to zoom in, as [x1, y1, x2, y2], where (x1, y1) is the top-left corner and (x2, y2) is the bottom-right corner. \par
- label (required string): The name or label of the object in the specified bounding box. \par
\end{prompttablebox}

\begin{prompttablebox}{Tool schema for text retrieval.}{tab:prompt_eval_text_tool}{Evaluation --- Text Search}
Name: Text Search \par
Implementation identifier: text\_search\_tool \par
Description: Search the web for text information related to your query. Use this when you need to find facts, news, or information. \par
\smallskip
Arguments: \par
- query (required string): The search query to find relevant information. \par
\end{prompttablebox}

\begin{prompttablebox}{Tool schema for reverse-image retrieval.}{tab:prompt_eval_image_tool}{Evaluation --- Image Search}
Name: Image Search \par
Implementation identifier: image\_search\_tool \par
Description: Perform a reverse image search to find similar images and related information. This can help identify objects, places, or find more context about the image. \par
\smallskip
Arguments: none. \par
\end{prompttablebox}

\subsection{Fully Expanded Search--Reasoning Example}
\label{appendix:expanded_search_example}

Figure~\ref{fig:evaluation_trajectory} compresses several intermediate turns
into the block labeled ``Iterative Search--Reasoning Loop.'' For completeness,
Table~\ref{tab:expanded_search_trajectory} expands the entire logical trace of
that example. The task asks for the difference between the founding years of
two cities reached from two shipping-company logos in the input image. The
first orange-container logo is grounded as Hapag-Lloyd and the second as OOCL.
The Search Agent must keep these anchors separate while traversing two evidence
chains and only then perform the requested subtraction.

\begin{table*}[t]
\centering
\scriptsize
\caption{Fully expanded trajectory corresponding to Figure~\ref{fig:evaluation_trajectory}. ``State update'' records the evidence retained for subsequent rounds; it is not an additional annotation exposed to the model.}
\label{tab:expanded_search_trajectory}
\setlength{\tabcolsep}{3.5pt}
\renewcommand{\arraystretch}{1.05}
\begin{tabularx}{\textwidth}{p{0.04\textwidth} p{0.14\textwidth} X X}
\toprule[1.2pt]
\textbf{Turn} & \textbf{Action} & \textbf{Search objective or observation} & \textbf{State update} \\
\midrule
1 & \textit{Image Search} & Reverse-search the full input image to identify the port scene and obtain visually similar results. & The returned titles and thumbnails locate the scene at the Port of Callao, Peru, but do not yet identify both required shipping companies. \\
\midrule
2 & \textit{Image Crop} & Inspect the upper-left orange container using a bounding box over its logo. & The crop exposes the text and livery of \textbf{Hapag-Lloyd}; store it as visual anchor $e_1$. \\
\midrule
3 & \textit{Image Crop} & Reinspect the \emph{original} image and crop a second orange container in the middle-right region. & The crop exposes \textbf{OOCL}; store it as a distinct visual anchor $e_2$. This second crop is taken from the original image, not from the first crop. \\
\midrule
4 & \textit{Text Search} & Query the history of Hapag-Lloyd and the shipping alliance to which it belonged. & Hapag-Lloyd is connected to \textbf{THE Alliance}, establishing the first relation after $e_1$. \\
\midrule
5 & \textit{Text Search} & Disambiguate the clue referring to a musical collective with the same name as the shipping alliance and identify its founder. & The name \textbf{THE Alliance} also refers to a Jamaican dancehall collective founded by \textbf{Bounty Killer}. The two same-name entities remain distinct nodes linked by the query's name-sharing clue. \\
\bottomrule[1.2pt]
\end{tabularx}
\end{table*}

\begin{table*}[t]
\centering
\scriptsize
\caption{Fully expanded trajectory corresponding to Figure~\ref{fig:evaluation_trajectory} (continued).}
\setlength{\tabcolsep}{3.5pt}
\renewcommand{\arraystretch}{1.05}
\begin{tabularx}{\textwidth}{p{0.04\textwidth} p{0.14\textwidth} X X}
\toprule[1.2pt]
\textbf{Turn} & \textbf{Action} & \textbf{Search objective or observation} & \textbf{State update} \\
\midrule
6 & \textit{Text Search} & Identify Bounty Killer's birthplace and the relevant Caribbean state. & Bounty Killer was born in \textbf{Jamaica}; Jamaica supplies the Caribbean-island-nation constraint in the question. \\
\midrule
7 & \textit{Text Search} & Find the regional intergovernmental agency of which Jamaica is a member and determine the city serving as its seat. & Jamaica is a member of \textbf{OPANAL}, whose seat is in \textbf{Mexico City}. A subsequent date lookup gives Mexico City's founding year as \textbf{1521}. \\
\midrule
8 & \textit{Text Search} & Starting again from $e_2$, determine the region in which OOCL is headquartered. & OOCL is headquartered in \textbf{Hong Kong}; store Hong Kong as the first nonvisual node of the second chain. \\
\midrule
9 & \textit{Text Search} & Follow Hong Kong's active membership in the global intergovernmental grouping and locate that grouping's secretariat. & Hong Kong participates in \textbf{APEC}, and the \textbf{APEC Secretariat} is located in \textbf{Singapore}. A date lookup gives Singapore's founding year as \textbf{1819}. \\
\midrule
10 & Reason and answer & Bind 1521 to the first chain and 1819 to the second chain, preserve the subtraction order specified by the question, and calculate $1819-1521$. & The Search Agent submits the final answer \textbf{298}. \\
\bottomrule[1.2pt]
\end{tabularx}
\end{table*}

The two recovered chains are therefore
\begin{align*}
e_1:\quad
&\text{Hapag-Lloyd}
\rightarrow \text{THE Alliance (shipping)}\\
&\rightarrow \text{THE Alliance (music)}
\rightarrow \text{Bounty Killer}\\
&\rightarrow \text{Jamaica}
\rightarrow \text{OPANAL}
\rightarrow \text{Mexico City}
\rightarrow 1521,\\[3pt]
e_2:\quad
&\text{OOCL}
\rightarrow \text{Hong Kong}
\rightarrow \text{APEC}\\
&\rightarrow \text{APEC Secretariat}
\rightarrow \text{Singapore}
\rightarrow 1819.
\end{align*}
Here, Mexico City is the seat of OPANAL; Jamaica, rather than Mexico City, is
the Caribbean island nation in the clue. Likewise, OOCL is headquartered in
Hong Kong; Singapore is reached downstream as the location of the APEC
Secretariat. Keeping these roles separate prevents the two relation chains from
being collapsed into incorrect direct claims. The requested result is
\begin{equation*}
1819-1521=\boxed{298}.
\end{equation*}

After the example-specific expansion above,
Algorithm~\ref{alg:tool_augmented_inference} gives the corresponding general
iterative search--reasoning procedure used for all evaluation instances.
The parser accepts at most one tool call from each model response. Every tool
result is wrapped in \texttt{<tool\_response>} tags and appended to the dialogue
history as a user message, allowing the Search Agent to condition its next
decision on all previously accumulated visual and textual evidence. Image Crop
always operates on the original image rather than the previous crop, which
allows the agent to revisit the visual input and inspect a different region.

\begin{algorithm*}[t]
\footnotesize
\caption{Expanded Iterative Search--Reasoning Loop in \eval}
\label{alg:tool_augmented_inference}
\KwIn{Image $I$, query $q$, base prompt $p$, tool definitions $\mathcal{T}$, maximum rounds $N$, text depth $K_t=3$, image depth $K_i=5$}
\KwOut{Final response $a$, tool-call trace $\tau$, termination state $s$}
$I' \leftarrow \operatorname{Preprocess}(I)$\;
$H \leftarrow [(\textsc{System},\,p\oplus\mathcal{T}),
(\textsc{User},\,(I',q))]$; $E\leftarrow\varnothing$; $\tau\leftarrow[\,]$\;
\For{$n\leftarrow 1$ \KwTo $N$}{
    Generate $y_n$ from the complete history $H$, including the query, images, thoughts, and accumulated evidence $E$\;
    \If{generation returns an error}{
        \KwRet{$(y_n,\tau,\textsc{Error})$}\;
    }
    Locate the first \texttt{<tool\_call>}\ldots\texttt{</tool\_call>} span in $y_n$\;
    \If{no such span exists}{
        $a\leftarrow y_n$ and stop without invoking another tool\;
        \KwRet{$(a,\tau,\textsc{AnswerSubmitted})$}\;
    }
    Decode the span as JSON to obtain tool identifier $u$ and arguments $\theta$\;
    \If{JSON decoding fails}{
        $a\leftarrow y_n$ and stop because no executable tool call was parsed\;
        \KwRet{$(a,\tau,\textsc{AnswerSubmitted})$}\;
    }
    Append $(\textsc{Assistant},y_n)$ to $H$\;
    \uIf{$u=\texttt{image\_zoom\_in\_tool}$ (\textit{Image Crop})}{
        Read $[x_1,y_1,x_2,y_2]$ from $\theta$ and verify that it contains four valid coordinates\;
        \eIf{the bounding box is valid}{
            Crop $I[y_1:y_2,x_1:x_2]$ from the original image $I$ and apply the model's image preprocessing\;
            $r_n\leftarrow$ ``Here is the zoomed image'' together with the processed crop\;
        }{
            $r_n\leftarrow$ invalid-bounding-box error\;
        }
    }
    \uElseIf{$u=\texttt{text\_search\_tool}$ (\textit{Text Search})}{
        Read search query $z$ from $\theta$ and verify that the search and summarization services are available\;
        \eIf{$z$ and the required services are valid}{
            Submit $z$ to the search backend and retain the top-$K_t$ organic results\;
            \ForEach{retained result $(t_j,\ell_j,s_j)$}{
                Fetch webpage $\ell_j$; skip inaccessible and unsupported resources\;
                Truncate the fetched text to 30,000 characters\;
                Use the webpage-summarization prompt to produce a summary of at most five sentences\;
            }
            Concatenate $z$, result titles, links, snippets, and all successful page summaries\;
            Apply the same summarization prompt once more to obtain the final search evidence $r_n$\;
        }{
            $r_n\leftarrow$ search-unavailable error\;
        }
    }
    \uElseIf{$u=\texttt{image\_search\_tool}$ (\textit{Image Search})}{
        Read the reverse-image-search record associated with $I$\;
        \eIf{titles and thumbnails are available}{
            Retain up to $K_i$ results and interleave each title with its processed thumbnail in $r_n$\;
        }{
            $r_n\leftarrow$ ``No matching images were found''\;
        }
    }
    \Else{
        $r_n\leftarrow$ unknown-tool error\;
    }
    Append $(u,\theta)$ to the tool-call trace $\tau$\;
    Wrap $r_n$ in \texttt{<tool\_response>}\ldots\texttt{</tool\_response>} tags\;
    Append the textual or multimodal evidence in $r_n$ to $E$\;
    Append $(\textsc{User},r_n)$ to $H$ for the next reasoning round\;
}
\KwRet{$(\varnothing,\tau,\textsc{MaxRounds})$}\;
\end{algorithm*}

\subsection{Webpage Summarization Prompts}

The following prompt pair is used both for individual webpage summaries and for
the final aggregation of retrieved content. For an individual page,
\{content\_limit\} is 30,000 characters.

\begin{prompttablebox}{System prompt used by the webpage summarizer.}{tab:prompt_eval_summary_system}{Evaluation --- Webpage summarizer system prompt}
You are a helpful assistant. Your task is to summarize the main content of the given web page in no more than five sentences. Your summary should cover the overall key points of the page, not just parts related to the user's question. \par
\smallskip
If any part of the content is helpful for answering the user's question, be sure to include it clearly in the summary. Do not ignore relevant information, but also make sure the general structure and main ideas of the page are preserved. Your summary should be concise, factual, and informative. \par
\end{prompttablebox}

\begin{prompttablebox}{User prompt used by the webpage summarizer.}{tab:prompt_eval_summary_user}{Evaluation --- Webpage summarizer user prompt}
Webpage Content (first \{content\_limit\} characters) is: \{content\} \par
Question: \{query\} \par
\end{prompttablebox}

\subsection{LLM-as-Judge Prompts}

\begin{prompttablebox}{System prompt used for automatic answer assessment.}{tab:prompt_eval_judge_system}{Evaluation --- LLM judge system prompt}
You are an AI assistant tasked with evaluating the correctness of model responses based on an image, question, and ground truth answer. Your judgment should follow these principles: \par
\smallskip
1. Consider the image, question, and ground truth answer holistically before evaluating the model's response. \par
2. Your decision should be strictly Yes or No, based on whether the model's response is factually accurate and aligns with the ground truth answer. \par
3. If the model response is a more specific form of the ground truth answer, it is correct. \par
4. If the model response includes all key information but adds minor details, it is correct as long as the extra details are factually correct. \par
5. If the model response contradicts, modifies, or omits critical parts of the answer, it is incorrect. \par
6. For numerical values, ensure correctness even when presented in different units. \par
7. For names, check for first and last name correctness. If the middle name is extra but correct, consider it correct. \par
8. For yes/no questions, the response must exactly match ``Yes'' or ``No'' to be correct. \par
9. If the judgment can be made based solely on the text, you may choose to ignore the input image, as some images may be unfamiliar to you and could affect your judgment. Refer to the image only when necessary to minimize misjudgment. \par
10. If there are multiple candidate answers, you can also evaluate the model's response against all of them. If the response aligns with at least one candidate according to the rules above, it should be considered correct. \par
11. For multiple choice questions (A, B, C, D), be more lenient. If the model provides the correct letter choice, even with additional text or formatting, consider it correct. \par
12. If the model's answer contains the correct choice letter (A, B, C, or D) anywhere in the response, and it's clear this is the intended answer, mark it as correct. \par
13. Ignore formatting issues like extra parentheses, brackets, or minor text variations as long as the core answer is correct. \par
\smallskip
Your output must be in the following format: \par
$<$judge$>$Yes/No$<$/judge$>$ \par
$<$reason$>$Explanation of why the answer is correct or incorrect.$<$/reason$>$ \par
\end{prompttablebox}

\begin{prompttablebox}{User prompt used for automatic answer assessment.}{tab:prompt_eval_judge_user}{Evaluation --- LLM judge user prompt}
Image, Question, and Model Response Evaluation \par
Question: \{question\} \par
Ground Truth Answer: \{ground\_truth\} \par
Model Response: \{model\_response\} \par
\smallskip
Evaluation Instructions \par
Evaluate whether the Model Response is correct based on the Image, Question and Ground Truth Answer. Follow the predefined judgment rules and provide a clear Yes/No answer along with a justification. \par
\smallskip
Output Format \par
$<$judge$>$Yes/No$<$/judge$>$ \par
$<$reason$>$Detailed reasoning following the evaluation principles.$<$/reason$>$ \par
\end{prompttablebox}

\end{appendices}

\end{document}